\pdfoutput=1
\documentclass[manuscript=article]{achemso}
\usepackage[table,dvipsnames]{xcolor}
\usepackage{tikz}
\usepackage{float}
\usepackage{amsmath}
\usepackage{amssymb}
\usepackage{siunitx}
\usepackage{booktabs}
\usepackage{geometry}
\usepackage{rotating}
\usepackage{multirow}
\usepackage{mathtools}
\usepackage{prettyref}
\usepackage{algorithm}
\usepackage{algpseudocode}
\usepackage{threeparttable}
\usepackage[utf8]{inputenc}
\usepackage[colorlinks,linkcolor=black,citecolor=blue]{hyperref}
\newrefformat{fig}{Figure~\ref{#1}}
\newrefformat{table}{Table~\ref{#1}}
\newrefformat{eq}{Equation~(\ref{#1})}
\newrefformat{alg}{Algorithm~\ref{#1}}
\newrefformat{sec}{Section~\ref{#1}}
\newcommand{\tabincell}[2]{\begin{tabular}{@{}#1@{}}#2\end{tabular}}
\definecolor{c00cc00}{RGB}{0,204,0}
\definecolor{ccccc00}{RGB}{204,204,0}
\definecolor{c33cccc}{RGB}{51,204,204}
\setkeys{acs}{keywords = true, abbreviations = true}

\author{Nianze TAO}
\email{tao-nianze@hiroshima-u.ac.jp}
\affiliation[Tokyo University of Agriculture and Technology]{Department of Applied Physics and Chemical Engineering, Faculty of Engineering, Tokyo University of Agriculture and Technology, 2-24-16 Naka-cho, Koganei-shi, Tokyo, Japan 184-8588}
\alsoaffiliation[Hiroshima University]
{Department of Chemistry, Graduate School of Advanced Science and Engineering, Hiroshima University, 1-3-1 Kagamiyama, Higashi-Hiroshima, Japan 739-8524}
\author{Minori ABE}
\email{minoria@go.tuat.ac.jp}
\affiliation[Tokyo University of Agriculture and Technology]{Department of Applied Physics and Chemical Engineering, Faculty of Engineering, Tokyo University of Agriculture and Technology, 2-24-16 Naka-cho, Koganei-shi, Tokyo, Japan 184-8588}

\title{Sampling Out-of-Distribution Chemical Spaces via Bayesian Flow}

\abbreviations{BFN: Bayesian Flow Network\\ OOD: out-of distribution\\ SOTA: state-of-the-art\\ DM: diffusion model\\ SDE: stochastic differential equation\\ ODE: ordinary differential equation\\ SAR: semi-autoregressive\\ FCD: Fr\'echet ChemNet Distance\\ QED: quantitave estimate drug-likeness\\ SA: synthetic accessibility\\ DS: docking score\\ SNN: similarity to the nearest neighbour\\ SASA: solvent accessible surface area\\ SMILES: Simplified Molecular Input Line Entry System\\ SELFIES: Self-Referencing Embedded Strings\\ RL: reinforcement learning\\ UMAP: Uniform Manifold Approximation and Projection\\ MDLM: masked diffusion language model}

\begin{tocentry}
	\includegraphics[width=8.47cm,height=4.67cm]{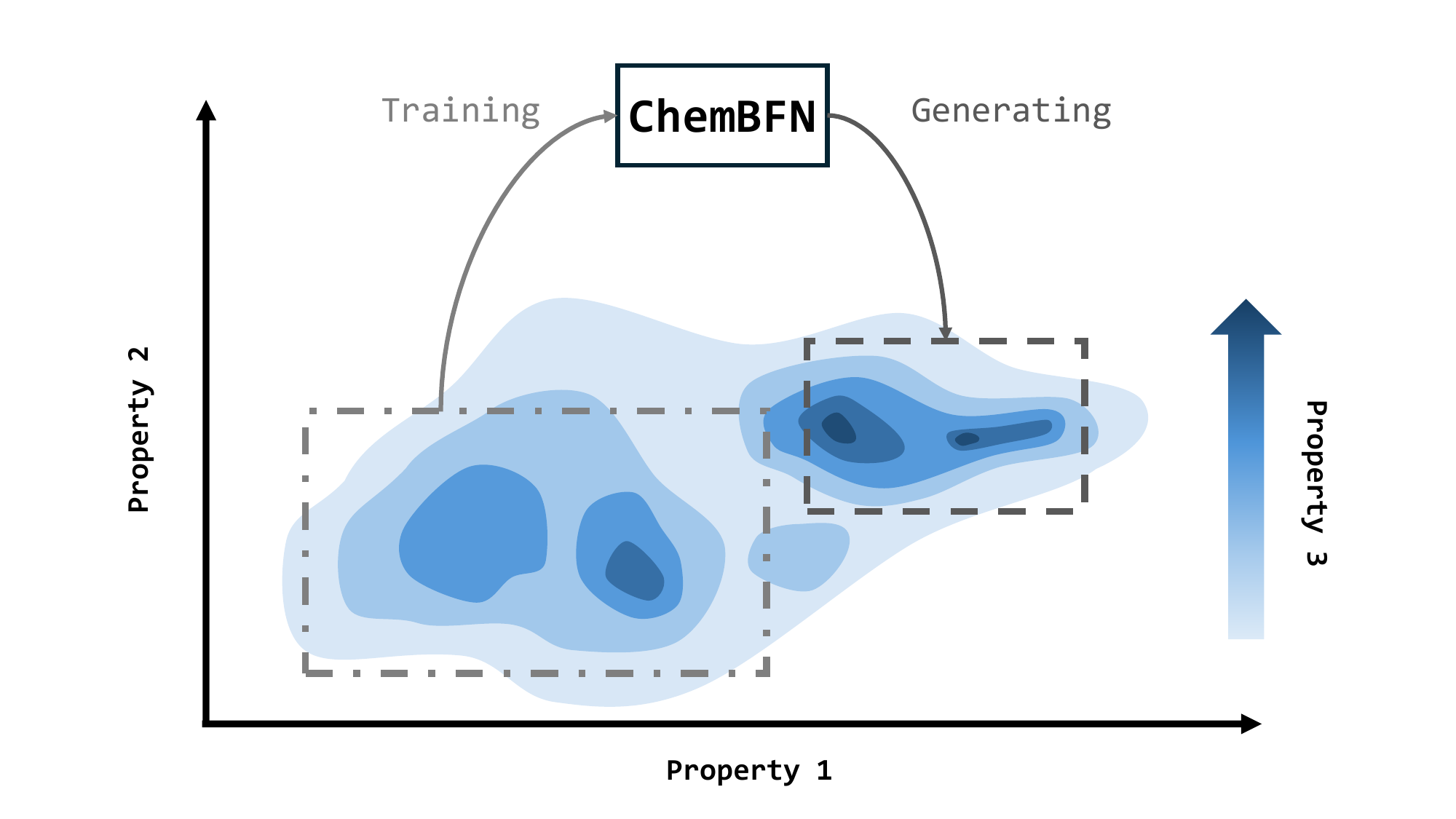}
\end{tocentry}

\begin{document}
	
	\begin{abstract}
		Generating novel molecules with higher properties than the training space, namely the out-of-distribution generation, is important for \textit{de novo} drug design. However, it is not easy for distribution learning-based models, for example diffusion models, to solve this challenge as these methods are designed to fit the distribution of training data as close as possible. In this paper, we show that Bayesian flow network, especially ChemBFN model, is capable of intrinsically generating high quality out-of-distribution samples that meet several scenarios. A reinforcement learning strategy is added to the ChemBFN and a controllable ordinary differential equation solver-like generating process is employed that accelerate the sampling processes. Most importantly, we introduce a semi-autoregressive strategy during training and inference that enhances the model performance and surpass the state-of-the-art models. A theoretical analysis of out-of-distribution generation in ChemBFN with semi-autoregressive approach is included as well.
	\end{abstract}
	
	\section*{Scientific Contribution}
	\par Benchmarked on both small molecule and protein generative tasks, ChemBFN method showed outstanding out-of-distribution performance without complex modification, proving its feasibility of exploring chemical spaces outside the training data and its value as a tool to accelerate drug design and materials discovery.
	
	\section{Introduction}
	\par {\LARGE T}he chemical space is large, and even the sub-space is too large to be explored completely. For instance, the number of \textit{small} drug-like molecules is believed to be over $10^{60}$\cite{drug-design-art,organic-substituents,chemical-space}, amongst which, however, only a much smaller sub-region, e.g., macrocycles\cite{mcr-drug,macrocycles-in-drug-discovery,exploration-of-macrocycles}, were tested in the laboratories and applied to real world challenges; when we start considering larger systems like proteins, the space grows exponentially\cite{protein-space}.
	\par Recent emerging deep generative models that virtually search the chemical space are attractive alternatives to trial-and-error processes conducted by human scientists\cite{smiles-vae,mood,conditional-vae,generative-models-for-automatic-chemical-design,deepicl}. Despite the success of generative models, especially the distributional learning models, in molecule generation, some researchers have pointed out the limitations of these methods: because the models and benchmarks were designed to optimise and test the in-distribution performance, i.e., how \textit{close} the generated molecules are to the training data, (1) the models were poor at generating highly novel samples with desired properties\cite{assesse-ai,mood}; (2) multi-objective optimisation was difficult\cite{mood}; (3) the sampling space could change to the false-positive region volatilely when an overconfident guidance appeared\cite{cgd}. To overcome these problems, it is important to improve the performance of out-of-distribution (OOD) generation, i.e., generating compounds with higher properties than that of molecules in the training dataset.
	\par Pioneering works have introduced a few methods to enhance the diffusion models' OOD performance including introducing a dedicatedly designed control method to steer towards the OOD space\cite{mood}, and utilising unlabelled data to expand the training domain and regularise the model\cite{cgd}. In this research, we, on the other hand, show that the Bayesian flow network, another branch of deep generative methods, is a \textit{natural} OOD sampler. Moreover, we demonstrate that by introducing a semi-autoregressive (SAR) behaviour our method outperformed several state-of-the-art (SOTA) models in OOD multi-objective optimisation tasks.
	
	\section{Methods}
	\label{sec:methods}
	\subsection{Bayesian Flow Networks}
	\par Similar to denoising diffusion models\cite{ddpm,ddim,diffusion_probabilistic,score-based,diffusion-schrodinger-bridge,shortcutmodels} (DMs) and more general flow matching methods\cite{flow-matching,flow-and-diffusion,ot-cfm,simplexflow,dirichlet-flow,fisher-flow,sfm,discrete-flow-matching,flowmol-ctmc}, Bayesian flow networks\cite{bfn} (BFNs) splits the generative process into a sequential steps. Some research has pointed out that the generative process of BFN is similar to and thus can be approximated as a reversed stochastic differential equation (SDE) that used in DMs though\cite{bfn-sde,protbfn,molcraft}, BFN method does not require to define a diffusion process nor learns the noise distribution: instead, BFN directly optimises the parameters of a distribution towards a more informative direction, which makes it applicable to continuous, discretised, and discrete data as the \textit{parameters} of any real-world distribution is continuous\cite{bfn}. The previous studies have already applied the idea of BFN to 3-dimensional molecular conformation generation\cite{geobfn,molcraft} (discretised for atom types and continuous cases for coordinates), text-based molecule generation\cite{chembfn} (discrete case for structural tokens), and protein sequence generation\cite{protbfn} (discrete case for amino acid types), which proved its capability of understanding the chemical space. 
	\par In this research, we employ ChemBFN\cite{chembfn}, a BFN model with encoder-only transformer\cite{attention} architecture handling the 1-dimensional molecular representation originally designed to generate SMILES\cite{smiles} and SELFIES\cite{selfies} strings, to study BFN's potentials of OOD sampling, i.e., generating samples with high properties out of a low-property training space. ChemBFN introduces a new accuracy schedule $\beta(t)$ for discrete BFN that reduces the reconstruction loss and improves the model performance\cite{chembfn}. However, limitations remained, e.g., to generate high quality samples, the number of sampling steps should be as large as 1000 in some cases\cite{chembfn} which limited the feasibility of generating large molecules; the distributional distance between the generated samples and the training data is large compared with baselines\cite{chembfn}. in the latter sections, we introduce our solution to solve the first limitation. We also realise that the second mentioned limitation can be transformed into an advantage in OOD generating. We show that a small change in the training or sampling process of ChemBFN can significantly enhance the OOD generative performance.
	
	\subsection{Efficient Sampling with BFN}
	\par Although ChemBFN succeeded in generating diverse molecules, the ratio of valid SMILES was low when fewer sampling steps were used\cite{chembfn}. To reduce the sampling steps while retaining the high validity ratio, here we propose two methods.
	\newline {\bf I. Auxiliary reinforcement learning term}. For any parameters of categorical distribution $\boldsymbol{\theta}$ fed into the neural network, we define the following reinforcement learning (RL) term, inspired by REINFORCE algorithm\cite{reinforce,policy-gradient}, added into the training loss
	\begin{equation}
		L^{RL} = \eta\mathbb{E}_{t\sim U(0,1)}\left(\boldsymbol{e}^{(k)}(\hat{\boldsymbol{\theta}};t)\cdot(1-\delta_{c}(\boldsymbol{e}(\hat{\boldsymbol{\theta}};t)))\right),
	\end{equation}
	where $\eta$ is a scaling constant, $k=argmax(\boldsymbol{e}(\hat{\boldsymbol{\theta}};t))_{dim=-1}$, $\boldsymbol{e}(\hat{\boldsymbol{\theta}};t)$ is the estimated categorical distribution (i.e., output distribution in terms of BFN) at time $t$ via the neural network, and $\delta_{c}(.) = 1$ when criterion $c$ is satisfied but 0 elsewhere. To increase valid ratio of generated molecules, a na{\"i}ve strategy is to raise the possibilities that at any time step $t$ the output distributions correspond to valid molecules. Therefore, we set the criterion $c$ as $c \vcentcolon = \{\boldsymbol{e}(\hat{\boldsymbol{\theta}};t)\mathrm{~corresponds~to~a~valid~molecule}\}$. In practice, we found that $\eta=0.01$ was optimal in most cases whereas $\eta<0.001$ or $\eta>0.1$ led to degraded performance.
	\newline {\bf II. ODE-like generating process}. The linkage between SDE and BFN generating process was first discovered by K. Xue \textit{et al}\cite{bfn-sde}, who stated that by directly operating in the latent space $\Omega_{z}$ instead of the distribution parameter space $\Omega_{\theta}$, an ordinary differential equation (ODE) solver or an SDE solver that updated the latent variable $\boldsymbol{z}_{i}=\boldsymbol{z}_{i-1}+\alpha_{i}(K\boldsymbol{e}(\hat{\boldsymbol{\theta}}_{i-1};t_{i-1})-1)+\sqrt{K\alpha_{i}}\boldsymbol{\epsilon}$ (where $\boldsymbol{z}\in\Omega_{z}$, $\boldsymbol{\theta}\in\Omega_{\theta}$, $\boldsymbol{\epsilon}\sim\mathcal{N}(\boldsymbol{0},\boldsymbol{I})$, $\alpha_{i}=\beta(t_{i})-\beta(t_{i-1})$, $\beta(t)$ is a monotonic accuracy schedule\cite{bfn,chembfn}, and $K$ is the number of categories) could accelerate the generating process. A simplified alternative method proposed by T. Atkinson \textit{et al}\cite{protbfn} and Y. Qu \textit{et al}\cite{molcraft} reduced the complexity of the algorithm, based on which we found that scaling the randomness by a temperature coefficient $\tau>0$ further benefited the valid ratio of generated objects. Our ODE-like sampling algorithm is shown in \prettyref{alg:ode-samplingh}.
	\par In the later text, we show that the combination of these two methods significantly speeds up the molecular generation.
	\begin{algorithm}
		\caption{ODE-like sampling algorithm}
		\label{alg:ode-samplingh}
		\begin{algorithmic}
			\Require $\tau>0$, conditioning $\boldsymbol{y}\in\mathbb{R}^{f}\cup\phi$, $n,K\in\mathbb{N}$, accuracy schedule $\beta(t)$
			\State $\boldsymbol{z}\leftarrow\boldsymbol{0}$
			\For{$i=1$ to $n$}
			\State $t\leftarrow(i-1)/n$
			\State $s\leftarrow t+1/n$
			\State $\boldsymbol{\theta}\leftarrow softmax(\boldsymbol{z})_{dim=-1}$
			\State $\boldsymbol{\epsilon}\sim\mathcal{N}(\boldsymbol{0},\boldsymbol{I})$
			\State $\boldsymbol{e}(\hat{\boldsymbol{\theta}};t)\leftarrow\mathrm{DISCRETE\_OUTPUT\_DISTRIBUTION}(\boldsymbol{\theta},t,\boldsymbol{y})$
			\State $\boldsymbol{z}\leftarrow\beta(s)(K\boldsymbol{e}(\hat{\boldsymbol{\theta}};t)-1)+\sqrt{K\beta(s)\tau}\boldsymbol{\epsilon}$
			\EndFor
			\State $\boldsymbol{\theta}\leftarrow softmax(\boldsymbol{z})_{dim=-1}$
			\State $\boldsymbol{e}(\hat{\boldsymbol{\theta}};1)\leftarrow\mathrm{DISCRETE\_OUTPUT\_DISTRIBUTION}(\boldsymbol{\theta},1,\boldsymbol{y})$
			\State {\bf return} $argmax(\boldsymbol{e}(\hat{\boldsymbol{\theta}};1))_{dim=-1}$
		\end{algorithmic}
	\end{algorithm}
	
	\subsection{SAR Training and Sampling}
	\label{sec:strategy}
	\par
\begin{figure}[H]
		\centering
		\includegraphics[width=0.8\linewidth]{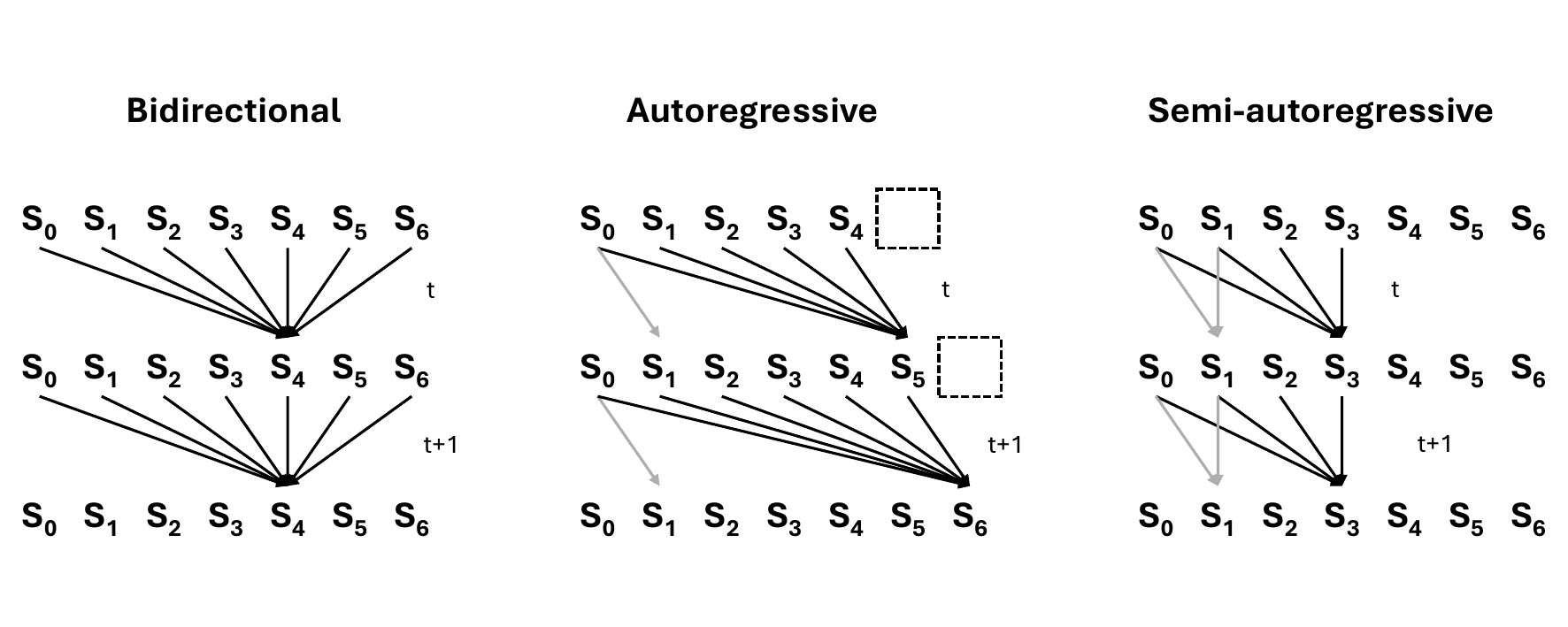}
		\caption{\small Visualisation of bidirectional, autoregressive, and SAR token update methods. Note that to make the equations clear, we use $t+1$, $i+1$, and $t_{i+1}$ in a interchangeable way in the main text when the discretised time is discussed.}
		\label{fig:strategy}
	\end{figure}

 In the original ChemBFN\cite{chembfn} model and any BERT\cite{bert}-like model, tokens are updated bidirectionally as illustrated in \prettyref{fig:strategy} (left), where for an arbitrary token in a finite fixed-length sequence $s_{j}\in(s_{0}, s_{1}, s_{2}, ..., s_{N})$ a function $f_{BI}$ maps $s_{j}$ from $i^{th}$ layer to $(i+1)^{th}$ layer based on itself and tokens from both left side (previous tokens) and right side (subsequent tokens), i.e., $s^{i+1}_{j}=f_{BI}(s^{i}_{j};s^{i}_{0:j-1},s^{i}_{j+1:N})$. In the case of autoregressive models, however, since the sequence is extended step by step (as shown in \prettyref{fig:strategy} (middle)), the next token $s_{j+1}$ only comes from the previous tokens via a function $f_{AR}$, i.e., $s^{i+1}_{j+1}=f_{AR}(s^{i}_{0:j})$. For models employing self-attention mechanism, e.g., the decoder of autoencoder transformer\cite{attention} model and GPT\cite{gpt} model, a causal mask that maps entities of the attention matrix above the main diagonal to zero is applied to implement this autoregressive behaviour. We found that in a trained ChemBFN model, the entries that were far away from the main diagonal in the attention matrices were extremely close to zero, which enlightened the possibility of applying causal masks to ChemBFN models without breaking models' generative capability. By employing the causal mask, we introduce a SAR behaviour to ChemBFN models, in which all tokens are updated together as a block but subsequent tokens are not used to update the current token (\prettyref{fig:strategy} (right)), i.e., $s^{i+1}_{j}=f_{SAR}(s^{i}_{j};s^{i}_{0:j-1})$.\par
\begin{figure}[H]
		\centering
		\includegraphics[width=1\linewidth]{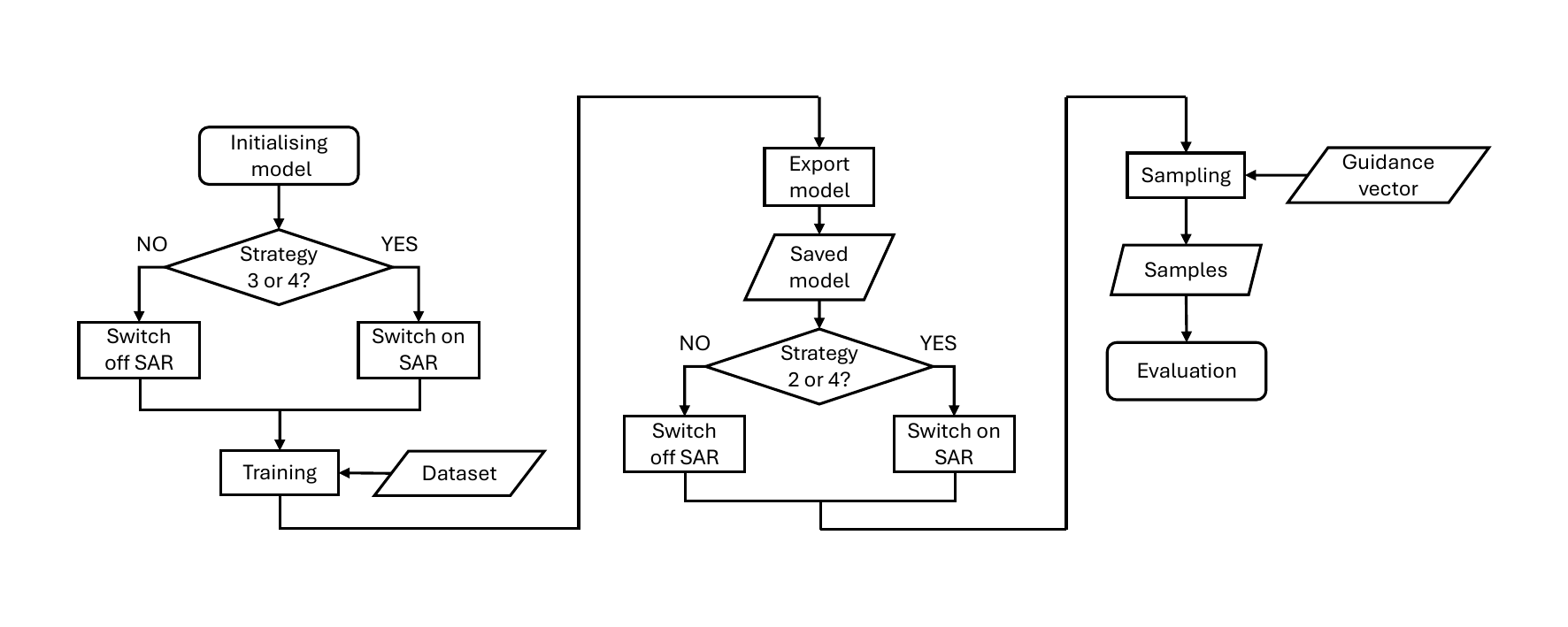}
		\caption{\small The flowchart showing that whether SAR is applied or not based on selected strategies listed in \prettyref{table:strategy}. `Switch on SAR' refers to applying causal masks to attention mechanism and `switch off SAR' refers to applying no causal masks. The guidance vector can be a null vector $\phi$ as described in the original paper of ChemBFN\cite{chembfn}.}
		\label{fig:strategy_workflow}
	\end{figure}

\begin{figure}[H]
		\centering
		\includegraphics[width=0.9\linewidth]{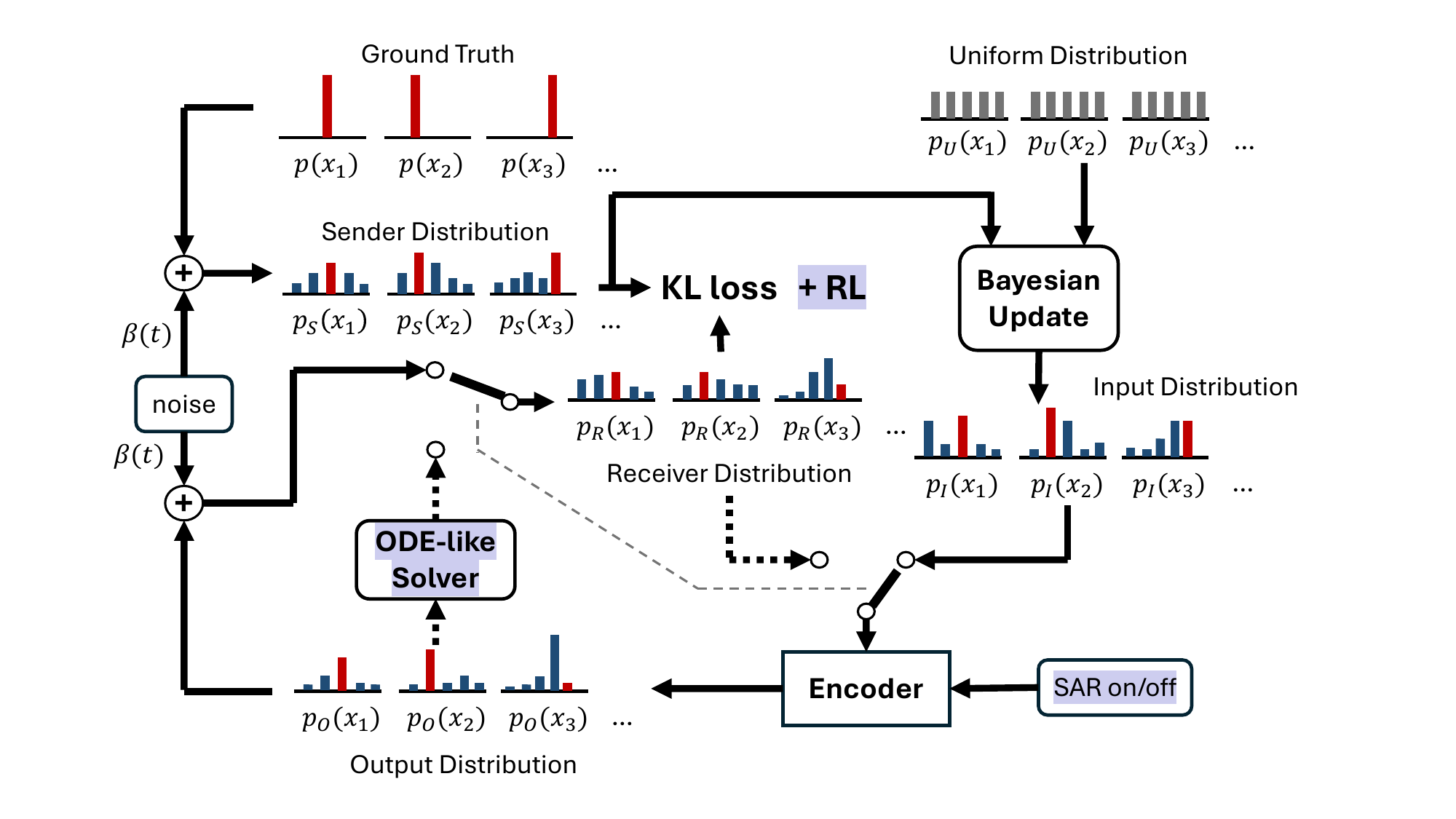}
		\caption{\small The overview schematic of integrating SAR, online RL, and ODE-like solver into ChemBFN model. Purple-highlighted parts are newly added methods in this work. The dashed path is the loop involved in inference process. The solid path leading to `KL loss + RL' strands for the training process.}
		\label{fig:chembfn_flowchart}
	\end{figure}

 To be simple, we denote applying causal masks to models as `SAR' while `normal' stands for disenabling causal masks. Depending on whether the causal masks are used in training or sampling processes, four strategies (strategy 1 - 4) are proposed in \prettyref{table:strategy} and a workflow is in \prettyref{fig:strategy_workflow}. An overview of integration of above mentioned methods are shown in \prettyref{fig:chembfn_flowchart}. We show how these settings affect the generative behaviours of our model in later texts.
	
	\begin{table}[H]
		\centering
		\sisetup{table-format=6.0}
		\setlength\cmidrulewidth{\heavyrulewidth}
		\caption{\small Training and sampling strategies\textsuperscript{\emph{a}}}
		\label{table:strategy}
		\begin{threeparttable}
		\begin{tabular}{clccc}
			\toprule
			{} & {Strategy} & {Training} & {Sampling} & {}\\
			\midrule
			{} & {1} & {normal} & {normal} & {}\\
			{} & {2} & {normal} & {SAR} & {}\\
			{} & {3} & {SAR} & {normal} & {}\\
			{} & {4} & {SAR} & {SAR} & {}\\
			\bottomrule
		\end{tabular}
		\begin{tablenotes}
			\item [a] SAR is `semi-autoregressive'.
		\end{tablenotes}
		\end{threeparttable}
	\end{table}
	
	\subsection{Datasets and Benchmarks}
	\par MOSES\cite{moses} as well as GuacaMol\cite{guacamol} are widely-used benchmarks in the studies of small molecule generation. Apart from testing the validity, uniqueness, diversity, and novelty of generated molecules, MOSES benchmark focuses more on the distribution learning performances of tested models, including Tanimoto similarity (SNN), fragment similarity (Frag), scaffold similarity (Scaf), and Fr\'echet ChemNet Distance\cite{fcd} (FCD). Although the purpose of MOSES is to estimate how \textit{close} the learnt distribution is to the training space, in this study it was utilised to showcase how \textit{far away} the generated space of our method could be from the training space while retaining the chemical meaningfulness.
	\par Lee \textit{et al}\cite{mood}, on the other hand, proposed ZINC250k dataset, which consists of 249,455 molecules collected from ZINC\cite{zinc} database and their paired quantitative estimate of drug-likeness\cite{qed} (QED), synthetic accessibility\cite{sa} (SA), and docking scores (DS) unit in kcal/mol to five different proteins (PARP1, FA7, 5HT1B, BRAF, and JAK2) calculated via QuickVina 2\cite{qvina2}, and the corresponding metrics to test the OOD multi-objective guided molecular generating performance. In this benchmark, a group of filters that should be applied to generated molecules are defined as	
	\begin{equation}
		\left\{\begin{array}{lcl}
			{\textrm{QED}} & {>} & {0.5}\\
			{\textrm{SA}} & {<} & {5}\\
			{\textrm{DS}} & {<} & {\rm \widetilde{DS}(\textrm{molecules in training data})}\\
			{\textrm{SNN}} & {<} & {0.4},\\
		\end{array}\right.
		\label{eq:filter}
	\end{equation}
	where $\widetilde{\rm DS}$ stands for median value of DS and SNN is the Morgan fingerprint Tanimoto similarity to the nearest neighbour in the training set. The two metrics, \textit{novel hit ratio} and \textit{novel top 5\% DS}, are therefore defined as	
	\begin{equation}
		\left\{\begin{array}{lcl}
			{\textrm{Novel hit ratio}} & {=} & {\dfrac{\textrm{\textnumero\;molecules passed all the filters}}{\textrm{\textnumero\;generated molecules}} \times 100\%}\\
			{\textrm{Novel top 5\% DS}} & {=} & {\overline{DS}(\textrm{top 5\% molecules passed all the filters})}.\\
		\end{array}\right.
		\label{eq:zinc250k_metrics}
	\end{equation}
	According to Lee \textit{et al}\cite{mood}, it is recommended to sample 3,000 molecules for each target protein and to repeat the experiment 5 times to obtain the mean and standard deviation of the metrics.
	\par Apart from studying the small molecule generating, we further employed a labelled protein dataset (90,990 sequences in the training set) published by N. Gruver \textit{et al}\cite{nos} to test our method on the generative tasks of larger systems. Amongst several structure-related labels, we selected the percentage of beta sheets and solvent accessible surface area (SASA) as the objective targets as suggested by N. Gruver \textit{et al}\cite{nos}.
	
	\section{Experiments and Results}	
	\par In this section we demonstrated how different training and sampling strategies (defined in the section of \nameref{sec:strategy}) affected generated spaces of unconditional cases, and then quantified the performance of our model in OOD multi-object optimisations. The improvement of sampling quality from our online RL method and ODE-like generating process was presented as well.
	\subsection{Fast Sampling}
	\begin{table}[H]
		\centering
		\sisetup{table-format=6.0}
		\setlength\cmidrulewidth{\heavyrulewidth}
		\caption{\small Metrics of MOSES and GuacaMol benchmarks when different numbers of sampling steps and methods were employed\textsuperscript{\emph{a}}}
		\label{table:metrics}
		\tiny
		\begin{threeparttable}
			\begin{tabular}{lc|ccccc}
				\toprule
				\multirow{2}*{\tabincell{c}{\scriptsize Methods\\}} & \multirow{2}*{\tabincell{c}{\scriptsize Step\\}} & \multicolumn{5}{c}{\scriptsize MOSES}\\
				{} & {} & {\scriptsize Valid $\uparrow$} & {\scriptsize Unique@1k $\uparrow$} & {\scriptsize Unique@10k $\uparrow$} & {\scriptsize Filter $\uparrow$} & {\scriptsize Novelty $\uparrow$}\\
				\midrule
				{\scriptsize ChemBFN\cite{chembfn}} & {1k} & {{\scriptsize 0.916} $\pm$ 0.001} & {{\scriptsize 1.0} $\pm$ 0.0} & {{\scriptsize 0.998} $\pm$ 0.000} & {{\scriptsize 0.987} $\pm$ 0.001} & {{\scriptsize 0.880} $\pm$ 0.002}\\
				{\scriptsize ChemBFN\cite{chembfn}} & {100} & {{\scriptsize 0.911} $\pm$ 0.002} & {{\scriptsize 1.0} $\pm$ 0.0} & {{\scriptsize 0.998} $\pm$ 0.000} & {{\scriptsize 0.985} $\pm$ 0.001} & {{\scriptsize 0.884} $\pm$ 0.002}\\
				{\scriptsize ChemBFN + RL} & {100} & {{\scriptsize 0.928} $\pm$ 0.002} & {{\scriptsize 1.0} $\pm$ 0.0} & {{\scriptsize 0.999} $\pm$ 0.000} & {{\scriptsize 0.987} $\pm$ 0.000} & {{\scriptsize 0.922} $\pm$ 0.001}\\
				{\scriptsize ChemBFN + ODE} & {100} & {{\scriptsize 1.0} $\pm$ 0.0} & {{\scriptsize 0.991} $\pm$ 0.005} & {{\scriptsize 0.910} $\pm$ 0.004} & {{\scriptsize 0.998} $\pm$ 0.000} & {{\scriptsize 0.571} $\pm$ 0.003}\\
				{\scriptsize ChemBFN + RL + ODE} & {100} & {{\scriptsize 0.999} $\pm$ 0.000} & {{\scriptsize 1.0} $\pm$ 0.0} & {{\scriptsize 0.992} $\pm$ 0.001} & {{\scriptsize 0.998} $\pm$ 0.000} & {{\scriptsize 0.797} $\pm$ 0.001}\\
				{\scriptsize ChemBFN\cite{chembfn}} & {10} & {{\scriptsize 0.835} $\pm$ 0.003} & {{\scriptsize 1.0} $\pm$ 0.0} & {{\scriptsize 0.999} $\pm$ 0.000} & {{\scriptsize 0.976} $\pm$ 0.001} & {{\scriptsize 0.921} $\pm$ 0.002}\\
				{\scriptsize ChemBFN + RL} & {10} & {{\scriptsize 0.852} $\pm$ 0.001} & {{\scriptsize 1.0} $\pm$ 0.0} & {{\scriptsize 0.999} $\pm$ 0.000} & {{\scriptsize 0.977} $\pm$ 0.000} & {{\scriptsize 0.946} $\pm$ 0.002}\\
				{\scriptsize ChemBFN + ODE} & {10} & {{\scriptsize 0.946} $\pm$ 0.000} & {{\scriptsize 1.0} $\pm$ 0.0} & {{\scriptsize 0.996} $\pm$ 0.001} & {{\scriptsize 0.990} $\pm$ 0.000} & {{\scriptsize 0.838} $\pm$ 0.003}\\
				{\scriptsize ChemBFN + RL + ODE} & {10} & {{\scriptsize 0.949} $\pm$ 0.001} & {{\scriptsize 1.0} $\pm$ 0.0} & {{\scriptsize 0.998} $\pm$ 0.000} & {{\scriptsize 0.992} $\pm$ 0.000} & {{\scriptsize 0.873} $\pm$ 0.002}\\
				\midrule
				\midrule
				{} & {} & \multicolumn{5}{c}{\scriptsize GuacaMol}\\
				{} & {} & \multicolumn{2}{c}{\scriptsize Valid $\uparrow$} & {\scriptsize Unique $\uparrow$} & \multicolumn{2}{c}{\scriptsize Novelty $\uparrow$}\\
				\midrule
				{\scriptsize ChemBFN\cite{chembfn}} & {1k} & \multicolumn{2}{c}{{\scriptsize 0.807} $\pm$ 0.003} & {{\scriptsize 0.818} $\pm$ 0.001} & \multicolumn{2}{c}{{\scriptsize 0.975} $\pm$ 0.001}\\
				{\scriptsize ChemBFN + RL} & {10} & \multicolumn{2}{c}{{\scriptsize 0.826} $\pm$ 0.003} & {{\scriptsize 0.821} $\pm$ 0.003} & \multicolumn{2}{c}{{\scriptsize 0.982} $\pm$ 0.000}\\
				{\scriptsize ChemBFN + ODE} & {10} & \multicolumn{2}{c}{{\scriptsize 0.879} $\pm$ 0.003} & {{\scriptsize 0.806} $\pm$ 0.005} & \multicolumn{2}{c}{{\scriptsize 0.939} $\pm$ 0.002}\\
				{\scriptsize ChemBFN + RL + ODE} & {10} & \multicolumn{2}{c}{{\scriptsize 0.863} $\pm$ 0.003} & {{\scriptsize 0.820} $\pm$ 0.002} & \multicolumn{2}{c}{{\scriptsize 0.980} $\pm$ 0.000}\\
				\bottomrule
			\end{tabular}
			\begin{tablenotes}
				\small
				\item [a] The training and sampling followed strategy 1. $\uparrow$ stands for the higher the better. $\tau=0.5$ for MOSES and $\tau=0.05$ for GuacaMol when ODE-like generative method was applied.
			\end{tablenotes}
		\end{threeparttable}
	\end{table}
	\par To compare with the baseline, i.e., the native ChemBFN model, we trained two ChemBFN models from scratch combined with the online RL described in \nameref{sec:methods} on MOSES and GuacaMol datasets respectively, then sampled the SMILES molecular representation with native BFN method and ODE-like approach. The main results were summarised in \prettyref{table:metrics}. We observed that (1) RL strategy marginally improved the validity and diversity (uniqueness and novelty) of generated samples; (2) ODE-like sampling strategy helped to drastically increase the validity in the cost of diversity; (3) the combination of RL and ODE-like generating process was the optimal strategy to outperform the baseline in validity while retaining the high diversity even when the number of sampling steps was reduced from 1k to 10. We also found that (not shown in \prettyref{table:metrics}) the temperature $\tau$ strongly controlled the trade-off between sample quality (validity) and diversity: a $\tau<0.01$ could lead to validity $\geq$ 99.5\% but the ratio of unique molecules $<$ 60\%; when $\tau=1$ the novelty was high but the validity was in the same level of native BFN sampling. In practice, $\tau=0.5$ is a value that balances validity and diversity when the sizes of molecules in the training data are highly homogeneous, otherwise the temperature needs reducing to as low as 0.02. Overall, ChemBFN + RL + ODE-like sampling strategy is capable of generating one valid SMILES string within 10.5 to 11.6 steps, which makes it possible to run on a laptop without GPUs. Note that when the number of sampling steps $n$ increases, because $\Delta t = 1/n$ decreases to match more to the continuous time during training, the performance will further improve in the cost of novelty. However, we acknowledge the limitation of auxiliary RL that requires 30\% more training time at most.
	
	\subsection{Unconditional Generation of Small Molecules}
	\par
\begin{figure}[H]
		\centering
		\includegraphics[width=0.9\linewidth]{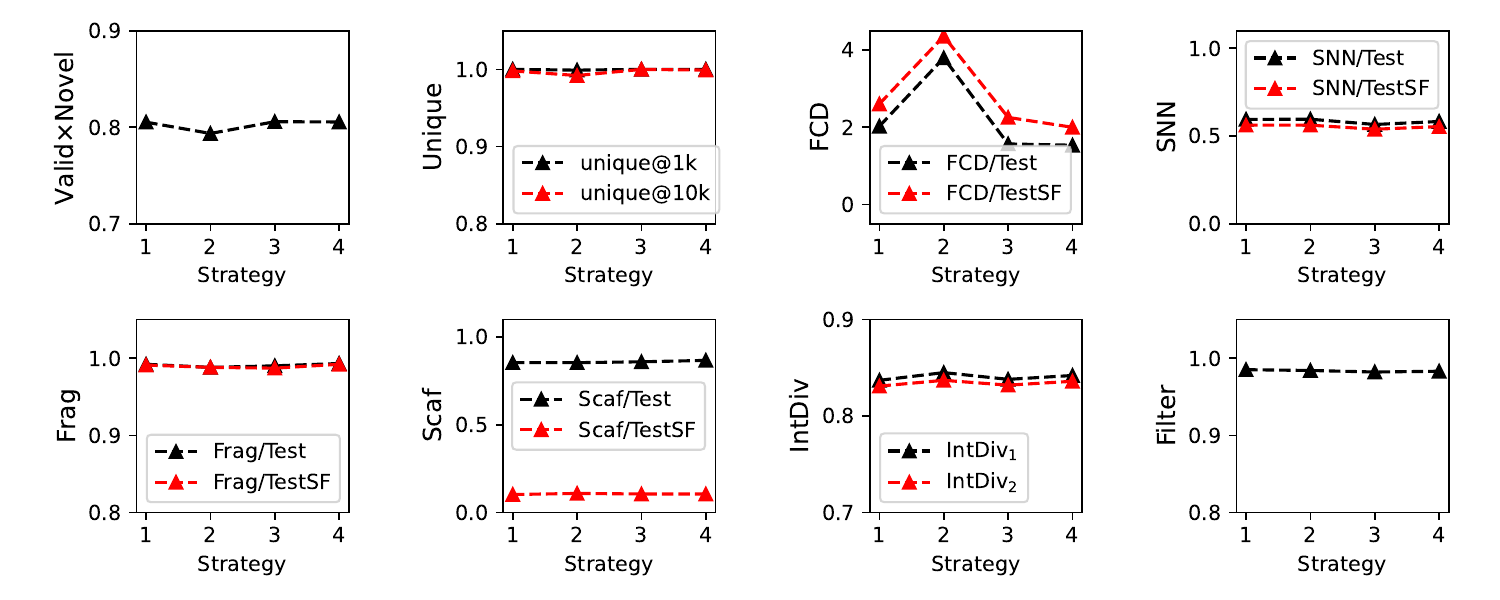}
		\caption{\small Visualisation of MOSES benchmark metrics of different strategies. `TestSF' was the scaffold-based test set. We reported Valid$\times$Novel values instead of validity and novelty separately. The strategy 1 - 4 are listed in \prettyref{table:strategy}.}
		\label{fig:moses_metrics}
	\end{figure}

\begin{figure}[H]
		\centering
		\includegraphics[width=0.9\linewidth]{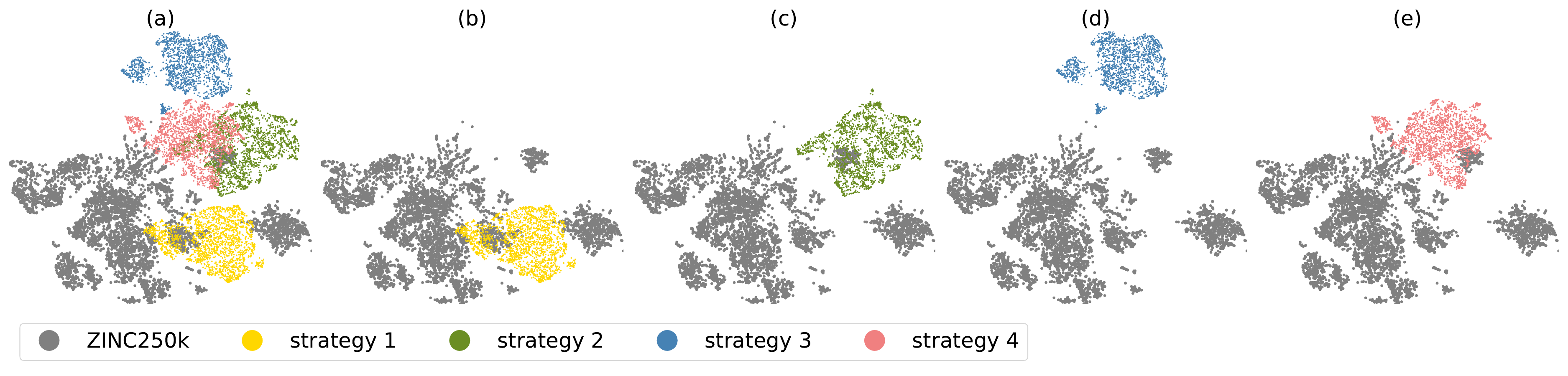}
		\caption{\small UMAP visualisation of the training space of ZINC250k dataset and the unconditionally generated sample spaces of different strategies. (a) the overview of chemical spaces explored by the different strategies compared with training data; (b) - (e) the chemical spaces sampled by strategies 1 to 4, respectively, with the background of training data.}
		\label{fig:zinc250k_uncond_umap}
	\end{figure}

 The MOSES testing metrics of ChemBFN using different strategies were visualised in \prettyref{fig:moses_metrics} and the 2-dimensional UMAP\cite{umap} (Uniform Manifold Approximation and Projection) plots of unconditional sample spaces of models trained on ZINC250k dataset against the training space were shown in \prettyref{fig:zinc250k_uncond_umap}. It is clear that diversity and structure related metrics did not significantly response to the change of training and/or sampling strategies, which indicated that SAR process did not worsen the model's capability of learning molecular structures. The key observation was that the OOD-ness indicated by the magnitude of FCD was strongly affected by different strategies. \prettyref{fig:zinc250k_uncond_umap} and \prettyref{fig:zinc250k_fcd} (a) further showed that the sample spaces were \textit{far away} from the training space while changing training and/or sampling strategies led the model to explore different OOD spaces.
	
	\subsection{Conditional Generation of Small Molecules}
	\par
\begin{figure}[H]
		\centering
		\includegraphics[width=0.9\linewidth]{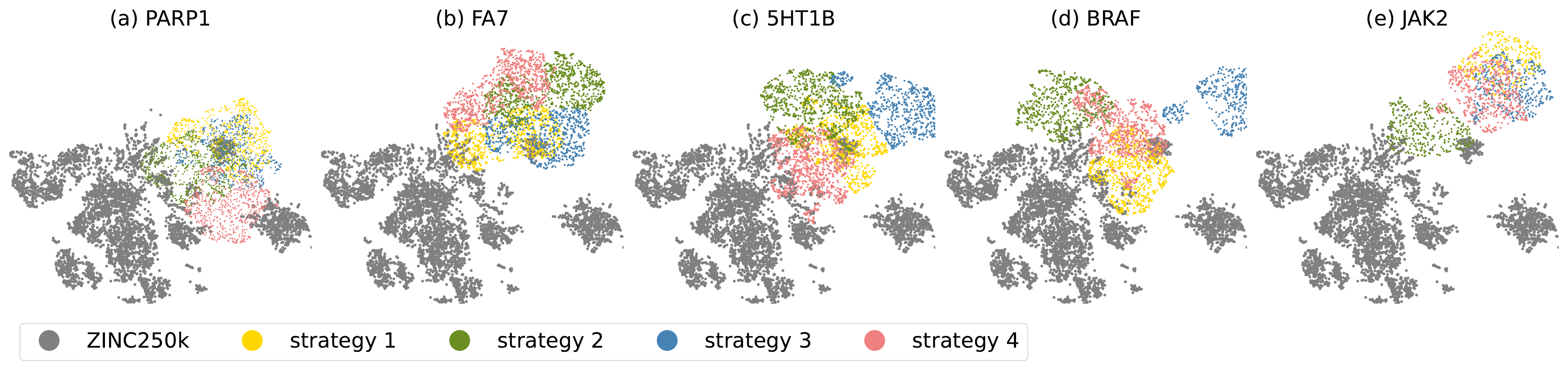}
		\caption{\small UMAP visualisation of the training space of ZINC250k dataset and the conditionally generated sample spaces of different strategies.}
		\label{fig:zinc250k_cond_umap}
	\end{figure}

\begin{figure}[H]
		\centering
		\includegraphics[width=0.9\linewidth]{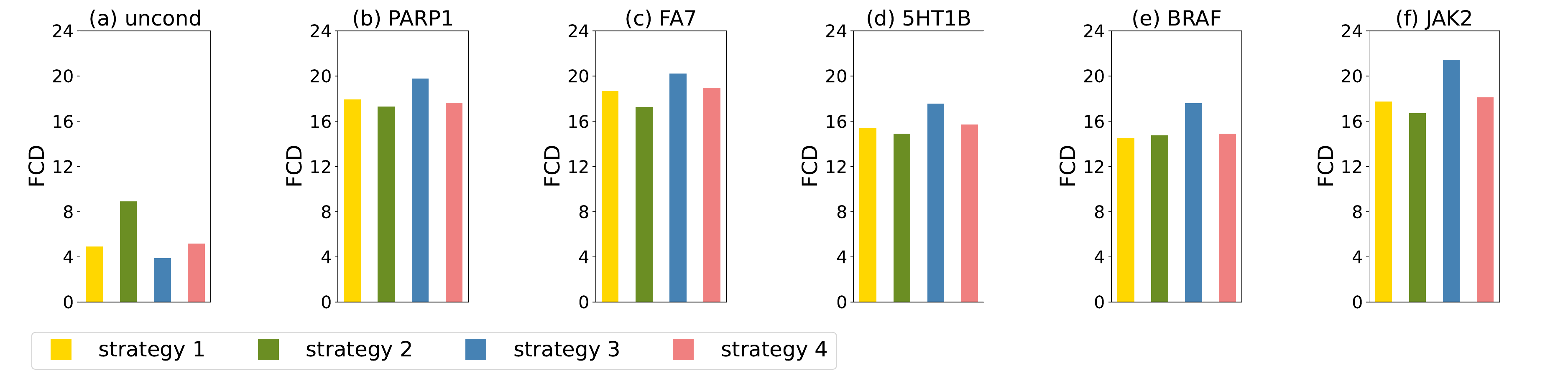}
		\caption{\small FCD values of unconditional and conditional samples of different strategies.}
		\label{fig:zinc250k_fcd}
	\end{figure}

 When a guidance vector $\textbf{y}=(\textrm{QED},\textrm{SA},\textrm{DS})$ pointing to a higher-property space than training space, i.e., high drug-likeness, low synthetic difficulty and more negative docking affinity, was applied to the sampling process via the classifier-free guidance method\cite{classifier-free}, we observed that the sample spaces had a tendency to be close to each other regardless the change of strategy (\prettyref{fig:zinc250k_cond_umap}). The OOD-ness was significantly larger than unconditional cases (\prettyref{fig:zinc250k_fcd}).
	
	\par In \prettyref{table:novel_hit_ratio_smiles} and \prettyref{table:novel_ds_smiles}, we summarised the novel hit ratios and novel top 5\% DS of ChemBFN when different training and/or sampling strategies were applied compared with the SOTA models\cite{reinvent,morld,hiervae,freed,gdss,mood} (note that the data of the SOTA models were provided by Lee \text{et al}\cite{mood}). Regarding the novel hit ratio, ChemBFN coupled with strategy 4 gave the best results on 4 out of 5 tasks amongst ChemBFN family, which, however, only outperformed the best method on 2 out of 5 tasks. Conversely, all our models outperformed all the SOTA methods for all 5 tasks when concerning the novel top 5\% docking scores, amongst which ChemBFN coupled with strategy 3 showed the best performance and strategy 4 presented the second best results.
	\begin{table}[H]
		\centering
		\sisetup{table-format=6.0}
		\setlength\cmidrulewidth{\heavyrulewidth}
		\caption{\small Novel hit ratios ($\uparrow$) compared with SOTA models\textsuperscript{\emph{a}}}
		\label{table:novel_hit_ratio_smiles}
		\begin{threeparttable}
		\tiny
		\begin{tabular}{lccccc}
			\toprule
			\multirow{2}*{\tabincell{c}{\small Methods\\}} & \multicolumn{5}{c}{Target proteins}\\
			{} & {PARP1} & {FA7} & {5HT1B} & {BRAF} & {JAK2}\\
			\midrule
			\midrule
			{\small REINVENT\cite{reinvent}} & {{\scriptsize 0.480} $\pm$ 0.344} & {{\scriptsize 0.213} $\pm$ 0.081} & {{\scriptsize 2.453} $\pm$ 0.561} & {{\scriptsize 0.127} $\pm$ 0.088} & {{\scriptsize 0.613} $\pm$ 0.167}\\
			{\small MORLD\cite{morld}} & {{\scriptsize 0.047} $\pm$ 0.050} & {{\scriptsize 0.007} $\pm$ 0.013} & {{\scriptsize 0.880} $\pm$ 0.735} & {{\scriptsize 0.047} $\pm$ 0.040} & {{\scriptsize 0.227} $\pm$ 0.118}\\
			{\small HierVAE\cite{hiervae}} & {{\scriptsize 0.553} $\pm$ 0.214} & {{\scriptsize 0.007} $\pm$ 0.013} & {{\scriptsize 0.507} $\pm$ 0.278} & {{\scriptsize 0.207} $\pm$ 0.220} & {{\scriptsize 0.227} $\pm$ 0.127}\\
			{\small FREED\cite{freed}} & {{\scriptsize 3.627} $\pm$ 0.961} & {{\scriptsize 1.107} $\pm$ 0.209} & {{\scriptsize\underline{10.187}} $\pm$ 3.306} & {{\scriptsize 2.067} $\pm$ 0.626} & {{\scriptsize 4.520} $\pm$ 0.673}\\
			{\small GDSS\cite{gdss}} & {{\scriptsize 1.933} $\pm$ 0.208} & {{\scriptsize 0.368} $\pm$ 0.103} & {{\scriptsize 4.667} $\pm$ 0.306} & {{\scriptsize 0.167} $\pm$ 0.134} & {{\scriptsize 1.167} $\pm$ 0.281}\\
			{\small MOOD\cite{mood}} & {{\scriptsize\textbf{7.017}} $\pm$ 0.428} & {{\scriptsize 0.733} $\pm$ 0.141} & {{\scriptsize\textbf{18.673}} $\pm$ 0.423} & {{\scriptsize 5.240} $\pm$ 0.285} & {{\scriptsize\textbf{9.200}} $\pm$ 0.524}\\
			\midrule
			{\small ChemBFN {\scriptsize strategy 1}} & {{\scriptsize 5.040} $\pm$ 0.473} & {{\scriptsize\underline{5.827}} $\pm$ 0.475} & {{\scriptsize 3.100} $\pm$ 0.264} & {{\scriptsize\textbf{5.340}} $\pm$ 0.341} & {{\scriptsize 3.973} $\pm$ 0.253}\\
			{\small ChemBFN {\scriptsize strategy 2}} & {{\scriptsize 4.660} $\pm$ 0.302} & {{\scriptsize 5.567} $\pm$ 0.153} & {{\scriptsize 3.653} $\pm$ 0.248} & {{\scriptsize\underline{5.293}} $\pm$ 0.529} & {{\scriptsize 4.040} $\pm$ 0.240}\\
			{\small ChemBFN {\scriptsize strategy 3}} & {{\scriptsize 4.287} $\pm$ 0.303} & {{\scriptsize 4.520} $\pm$ 0.383} & {{\scriptsize 3.860} $\pm$ 0.089} & {{\scriptsize 3.160} $\pm$ 0.314} & {{\scriptsize 4.093} $\pm$ 0.368}\\
			{\small ChemBFN {\scriptsize strategy 4}} & {{\scriptsize\underline{5.593}} $\pm$ 0.417} & {{\scriptsize\textbf{5.853}} $\pm$ 0.423} & {{\scriptsize 4.587} $\pm$ 0.358} & {{\scriptsize 4.233} $\pm$ 0.518} & {{\scriptsize\underline{5.260}} $\pm$ 0.314}\\
			\bottomrule
		\end{tabular}
		\begin{tablenotes}
			\small
			\item [a] The best results are in \textbf{\textit{bold}} and the second best results are \underline{underlined}. $\uparrow$ stands for the higher the better. We sampled the molecules for 1,000 steps in \textbf{SMILES} format. The results were the average values of 5 runs with the standard deviation reported.
		\end{tablenotes}
		\end{threeparttable}
	\end{table}
	
	\begin{table}[H]
		\centering
		\sisetup{table-format=6.0}
		\setlength\cmidrulewidth{\heavyrulewidth}
		\caption{\small Novel top 5\% DS ($\downarrow$) compared with SOTA models\textsuperscript{\emph{a}}}
		\label{table:novel_ds_smiles}
		\begin{threeparttable}
		\tiny
		\begin{tabular}{lccccc}
			\toprule
			\multirow{2}*{\tabincell{c}{\small Methods\\}} & \multicolumn{5}{c}{Target proteins}\\
			{} & {PARP1} & {FA7} & {5HT1B} & {BRAF} & {JAK2}\\
			\midrule
			\midrule
			{\small REINVENT\cite{reinvent}} & {{\scriptsize -8.702} $\pm$ 0.523} & {{\scriptsize -7.205} $\pm$ 0.264} & {{\scriptsize -8.770} $\pm$ 0.316} & {{\scriptsize -8.392} $\pm$ 0.400} & {{\scriptsize -8.165} $\pm$ 0.277}\\
			{\small MORLD\cite{morld}} & {{\scriptsize -7.532} $\pm$ 0.260} & {{\scriptsize -6.263} $\pm$ 0.165} & {{\scriptsize -7.869} $\pm$ 0.650} & {{\scriptsize -8.040} $\pm$ 0.337} & {{\scriptsize -7.816} $\pm$ 0.133}\\
			{\small HierVAE\cite{hiervae}} & {{\scriptsize -9.487} $\pm$ 0.278} & {{\scriptsize -6.812} $\pm$ 0.274} & {{\scriptsize -8.081} $\pm$ 0.252} & {{\scriptsize -8.978} $\pm$ 0.525} & {{\scriptsize -8.285} $\pm$ 0.370}\\
			{\small FREED\cite{freed}} & {{\scriptsize -10.427} $\pm$ 0.177} & {{\scriptsize -8.297} $\pm$ 0.094} & {{\scriptsize -10.425} $\pm$ 0.331} & {{\scriptsize -10.325} $\pm$ 0.164} & {{\scriptsize -9.624} $\pm$ 0.102}\\
			{\small GDSS\cite{gdss}} & {{\scriptsize -9.967} $\pm$ 0.028} & {{\scriptsize -7.775} $\pm$ 0.039} & {{\scriptsize -9.459} $\pm$ 0.101} & {{\scriptsize -9.224} $\pm$ 0.068} & {{\scriptsize -8.926} $\pm$ 0.089}\\
			{\small MOOD\cite{mood}} & {{\scriptsize -10.865} $\pm$ 0.113} & {{\scriptsize -8.160} $\pm$ 0.071} & {{\scriptsize -11.145} $\pm$ 0.042} & {{\scriptsize -11.063} $\pm$ 0.034} & {{\scriptsize -10.147} $\pm$ 0.060}\\
			\midrule
			{\small ChemBFN {\scriptsize strategy 1}} & {{\scriptsize\underline{-12.932}} $\pm$ 0.159} & {{\scriptsize -9.186} $\pm$ 0.149} & {{\scriptsize\textbf{-12.493}} $\pm$ 0.313} & {{\scriptsize -11.955} $\pm$ 0.143} & {{\scriptsize -11.792} $\pm$ 0.337}\\
			{\small ChemBFN {\scriptsize strategy 2}} & {{\scriptsize -12.528} $\pm$ 0.205} & {{\scriptsize -8.925} $\pm$ 0.083} & {{\scriptsize -11.912} $\pm$ 0.265} & {{\scriptsize -11.728} $\pm$ 0.103} & {{\scriptsize -11.677} $\pm$ 0.169}\\
			{\small ChemBFN {\scriptsize strategy 3}} & {{\scriptsize\textbf{-13.040}} $\pm$ 0.211} & {{\scriptsize\textbf{-9.611}} $\pm$ 0.154} & {{\scriptsize\underline{-12.448}} $\pm$ 0.241} & {{\scriptsize\textbf{-12.350}} $\pm$ 0.641} & {{\scriptsize\textbf{-12.111}} $\pm$ 0.275}\\
			{\small ChemBFN {\scriptsize strategy 4}} & {{\scriptsize -12.741} $\pm$ 0.205} & {{\scriptsize\underline{-9.442}} $\pm$ 0.074} & {{\scriptsize -12.275} $\pm$ 0.120} & {{\scriptsize\underline{-11.969}} $\pm$ 0.203} & {{\scriptsize\underline{-12.022}} $\pm$ 0.120}\\
			\bottomrule
		\end{tabular}
		\begin{tablenotes}
			\small
			\item [a] $\downarrow$ stands for the lower the better while other settings are the same as \prettyref{table:novel_hit_ratio_smiles}
		\end{tablenotes}
		\end{threeparttable}
	\end{table}
	
	\par We found that when a guidance vector pointing to a higher-property space than training space is applied, the ratio of invalid SMILES\cite{smiles} strings generated by the models increased drastically. To reduce the influence of hallucinations, we additionally trained two models: ChemBFN + RL + ODE and ChemBFN + SELFIES\cite{selfies}. The results were shown in \prettyref{table:novel_hit_ratio_smiles_2}, \prettyref{table:novel_ds_smiles_2}, \prettyref{table:novel_hit_ratio_selfies} and \prettyref{table:novel_ds_selfies}. The novel hit ratio for all target proteins, as shown in \prettyref{table:novel_hit_ratio_smiles_2} and \prettyref{table:novel_hit_ratio_selfies}, increased in both cases: ChemBFN + RL + ODE surpassed SOTA models in 4 out of 5 tasks even when the number of sampling steps was reduced from 1k to 100; for ChemBFN + SELFIES, the ratio significantly improved from less than 6\% to over 25\%, thereby outperforming all the SOTA models. The novel top 5\% docking scores dropped slightly as shown in \prettyref{table:novel_ds_smiles_2} and \prettyref{table:novel_ds_selfies}. Nevertheless, our models still surpassed SOTA methods.
	\begin{table}[H]
		\centering
		\sisetup{table-format=6.0}
		\setlength\cmidrulewidth{\heavyrulewidth}
		\caption{\small Novel hit ratios ($\uparrow$) of ChemBFN + RL + ODE-like sampling ($\tau=0.5$) compared with SOTA models\textsuperscript{\emph{a}}}
		\label{table:novel_hit_ratio_smiles_2}
		\begin{threeparttable}
			\tiny
			\begin{tabular}{lccccc}
				\toprule
				\multirow{2}*{\tabincell{c}{\small Methods\\}} & \multicolumn{5}{c}{Target proteins}\\
				{} & {PARP1} & {FA7} & {5HT1B} & {BRAF} & {JAK2}\\
				\midrule
				\midrule
				{\small REINVENT\cite{reinvent}} & {{\scriptsize 0.480} $\pm$ 0.344} & {{\scriptsize 0.213} $\pm$ 0.081} & {{\scriptsize 2.453} $\pm$ 0.561} & {{\scriptsize 0.127} $\pm$ 0.088} & {{\scriptsize 0.613} $\pm$ 0.167}\\
				{\small MORLD\cite{morld}} & {{\scriptsize 0.047} $\pm$ 0.050} & {{\scriptsize 0.007} $\pm$ 0.013} & {{\scriptsize 0.880} $\pm$ 0.735} & {{\scriptsize 0.047} $\pm$ 0.040} & {{\scriptsize 0.227} $\pm$ 0.118}\\
				{\small HierVAE\cite{hiervae}} & {{\scriptsize 0.553} $\pm$ 0.214} & {{\scriptsize 0.007} $\pm$ 0.013} & {{\scriptsize 0.507} $\pm$ 0.278} & {{\scriptsize 0.207} $\pm$ 0.220} & {{\scriptsize 0.227} $\pm$ 0.127}\\
				{\small FREED\cite{freed}} & {{\scriptsize 3.627} $\pm$ 0.961} & {{\scriptsize 1.107} $\pm$ 0.209} & {{\scriptsize 10.187} $\pm$ 3.306} & {{\scriptsize 2.067} $\pm$ 0.626} & {{\scriptsize 4.520} $\pm$ 0.673}\\
				{\small GDSS\cite{gdss}} & {{\scriptsize 1.933} $\pm$ 0.208} & {{\scriptsize 0.368} $\pm$ 0.103} & {{\scriptsize 4.667} $\pm$ 0.306} & {{\scriptsize 0.167} $\pm$ 0.134} & {{\scriptsize 1.167} $\pm$ 0.281}\\
				{\small MOOD\cite{mood}} & {{\scriptsize\underline{7.017}} $\pm$ 0.428} & {{\scriptsize 0.733} $\pm$ 0.141} & {{\scriptsize\textbf{18.673}} $\pm$ 0.423} & {{\scriptsize\underline{5.240}} $\pm$ 0.285} & {{\scriptsize\underline{9.200}} $\pm$ 0.524}\\
				\midrule
				{\small ChemBFN {\scriptsize strategy 3}} & {{\scriptsize 4.893} $\pm$ 0.596} & {{\scriptsize\underline{5.426}} $\pm$ 0.542} & {{\scriptsize 5.960} $\pm$ 0.505} & {{\scriptsize 4.740} $\pm$ 0.329} & {{\scriptsize 5.367} $\pm$ 0.426}\\
				{\small ChemBFN {\scriptsize strategy 4}} & {{\scriptsize\textbf{8.980}} $\pm$ 0.186} & {{\scriptsize\textbf{5.993}} $\pm$ 0.191} & {{\scriptsize\underline{11.940}} $\pm$ 0.638} & {{\scriptsize\textbf{12.573}} $\pm$ 0.569} & {{\scriptsize\textbf{9.827}} $\pm$ 0.570}\\
				\bottomrule
			\end{tabular}
			\begin{tablenotes}
				\small
				\item [a] The best results are in \textbf{\textit{bold}} and the second best results are \underline{underlined}. $\uparrow$ stands for the higher the better. We sampled the molecules with ODE-like generating process for 100 steps in \textbf{SMILES} format. The results were the average values of 5 runs with the standard deviation reported.
			\end{tablenotes}
		\end{threeparttable}
	\end{table}
	
	\begin{table}[H]
		\centering
		\sisetup{table-format=6.0}
		\setlength\cmidrulewidth{\heavyrulewidth}
		\caption{\small Novel top 5\% DS ($\downarrow$) of ChemBFN + RL + ODE-like sampling ($\tau=0.5$) compared with SOTA models\textsuperscript{\emph{a}}}
		\label{table:novel_ds_smiles_2}
		\begin{threeparttable}
			\tiny
			\begin{tabular}{lccccc}
				\toprule
				\multirow{2}*{\tabincell{c}{\small Methods\\}} & \multicolumn{5}{c}{Target proteins}\\
				{} & {PARP1} & {FA7} & {5HT1B} & {BRAF} & {JAK2}\\
				\midrule
				\midrule
				{\small REINVENT\cite{reinvent}} & {{\scriptsize -8.702} $\pm$ 0.523} & {{\scriptsize -7.205} $\pm$ 0.264} & {{\scriptsize -8.770} $\pm$ 0.316} & {{\scriptsize -8.392} $\pm$ 0.400} & {{\scriptsize -8.165} $\pm$ 0.277}\\
				{\small MORLD\cite{morld}} & {{\scriptsize -7.532} $\pm$ 0.260} & {{\scriptsize -6.263} $\pm$ 0.165} & {{\scriptsize -7.869} $\pm$ 0.650} & {{\scriptsize -8.040} $\pm$ 0.337} & {{\scriptsize -7.816} $\pm$ 0.133}\\
				{\small HierVAE\cite{hiervae}} & {{\scriptsize -9.487} $\pm$ 0.278} & {{\scriptsize -6.812} $\pm$ 0.274} & {{\scriptsize -8.081} $\pm$ 0.252} & {{\scriptsize -8.978} $\pm$ 0.525} & {{\scriptsize -8.285} $\pm$ 0.370}\\
				{\small FREED\cite{freed}} & {{\scriptsize -10.427} $\pm$ 0.177} & {{\scriptsize -8.297} $\pm$ 0.094} & {{\scriptsize -10.425} $\pm$ 0.331} & {{\scriptsize -10.325} $\pm$ 0.164} & {{\scriptsize -9.624} $\pm$ 0.102}\\
				{\small GDSS\cite{gdss}} & {{\scriptsize -9.967} $\pm$ 0.028} & {{\scriptsize -7.775} $\pm$ 0.039} & {{\scriptsize -9.459} $\pm$ 0.101} & {{\scriptsize -9.224} $\pm$ 0.068} & {{\scriptsize -8.926} $\pm$ 0.089}\\
				{\small MOOD\cite{mood}} & {{\scriptsize -10.865} $\pm$ 0.113} & {{\scriptsize -8.160} $\pm$ 0.071} & {{\scriptsize -11.145} $\pm$ 0.042} & {{\scriptsize -11.063} $\pm$ 0.034} & {{\scriptsize -10.147} $\pm$ 0.060}\\
				\midrule
				{\small ChemBFN {\scriptsize strategy 3}} & {{\scriptsize\textbf{-12.533}} $\pm$ 0.200} & {{\scriptsize\textbf{-9.383}} $\pm$ 0.093} & {{\scriptsize\textbf{-12.518}} $\pm$ 0.162} & {{\scriptsize\textbf{-12.019}} $\pm$ 0.170} & {{\scriptsize\underline{-11.861}} $\pm$ 0.190}\\
				{\small ChemBFN {\scriptsize strategy 4}} & {{\scriptsize\underline{-12.092}} $\pm$ 0.120} & {{\scriptsize\underline{-8.929}} $\pm$ 0.097} & {{\scriptsize\underline{-11.949}} $\pm$ 0.093} & {{\scriptsize\underline{-11.715}} $\pm$ 0.075} & {{\scriptsize\textbf{-12.181}} $\pm$ 0.198}\\
				\bottomrule
			\end{tabular}
			\begin{tablenotes}
				\small
				\item [a] $\downarrow$ stands for the lower the better while other settings are the same as \prettyref{table:novel_hit_ratio_smiles_2}
			\end{tablenotes}
		\end{threeparttable}
	\end{table}
	
	\begin{table}[H]
		\centering
		\sisetup{table-format=6.0}
		\setlength\cmidrulewidth{\heavyrulewidth}
		\caption{\small Novel hit ratios ($\uparrow$) compared with SOTA models\textsuperscript{\emph{a}}}
		\label{table:novel_hit_ratio_selfies}
		\begin{threeparttable}
		\tiny
		\begin{tabular}{lccccc}
			\toprule
			\multirow{2}*{\tabincell{c}{\small Methods\\}} & \multicolumn{5}{c}{Target proteins}\\
			{} & {PARP1} & {FA7} & {5HT1B} & {BRAF} & {JAK2}\\
			\midrule
			\midrule
			{\small REINVENT\cite{reinvent}} & {{\scriptsize 0.480} $\pm$ 0.344} & {{\scriptsize 0.213} $\pm$ 0.081} & {{\scriptsize 2.453} $\pm$ 0.561} & {{\scriptsize 0.127} $\pm$ 0.088} & {{\scriptsize 0.613} $\pm$ 0.167}\\
			{\small MORLD\cite{morld}} & {{\scriptsize 0.047} $\pm$ 0.050} & {{\scriptsize 0.007} $\pm$ 0.013} & {{\scriptsize 0.880} $\pm$ 0.735} & {{\scriptsize 0.047} $\pm$ 0.040} & {{\scriptsize 0.227} $\pm$ 0.118}\\
			{\small HierVAE\cite{hiervae}} & {{\scriptsize 0.553} $\pm$ 0.214} & {{\scriptsize 0.007} $\pm$ 0.013} & {{\scriptsize 0.507} $\pm$ 0.278} & {{\scriptsize 0.207} $\pm$ 0.220} & {{\scriptsize 0.227} $\pm$ 0.127}\\
			{\small FREED\cite{freed}} & {{\scriptsize 3.627} $\pm$ 0.961} & {{\scriptsize 1.107} $\pm$ 0.209} & {{\scriptsize 10.187} $\pm$ 3.306} & {{\scriptsize 2.067} $\pm$ 0.626} & {{\scriptsize 4.520} $\pm$ 0.673}\\
			{\small GDSS\cite{gdss}} & {{\scriptsize 1.933} $\pm$ 0.208} & {{\scriptsize 0.368} $\pm$ 0.103} & {{\scriptsize 4.667} $\pm$ 0.306} & {{\scriptsize 0.167} $\pm$ 0.134} & {{\scriptsize 1.167} $\pm$ 0.281}\\
			{\small MOOD\cite{mood}} & {{\scriptsize 7.017} $\pm$ 0.428} & {{\scriptsize 0.733} $\pm$ 0.141} & {{\scriptsize 18.673} $\pm$ 0.423} & {{\scriptsize 5.240} $\pm$ 0.285} & {{\scriptsize 9.200} $\pm$ 0.524}\\
			\midrule
			{\small ChemBFN {\scriptsize strategy 3}} & {{\scriptsize\underline{31.547}} $\pm$ 0.730} & {{\scriptsize\textbf{31.460}} $\pm$ 0.858} & {{\scriptsize\underline{25.047}} $\pm$ 0.837} & {{\scriptsize\underline{26.893}} $\pm$ 1.022} & {{\scriptsize\underline{34.060}} $\pm$ 0.710}\\
			{\small ChemBFN {\scriptsize strategy 4}} & {{\scriptsize\textbf{33.240}} $\pm$ 1.018} & {{\scriptsize\underline{30.133}} $\pm$ 0.442} & {{\scriptsize\textbf{30.933}} $\pm$ 0.573} & {{\scriptsize\textbf{29.033}} $\pm$ 0.647} & {{\scriptsize\textbf{42.133}} $\pm$ 1.509}\\
			\bottomrule
		\end{tabular}
		\begin{tablenotes}
			\small
			\item [a] The best results are in \textbf{\textit{bold}} and the second best results are \underline{underlined}. $\uparrow$ stands for the higher the better. We sampled the molecules for 1,000 steps in \textbf{SELFIES} format. The results were the average values of 5 runs with the standard deviation reported.
		\end{tablenotes}
		\end{threeparttable}
	\end{table}
	
	\begin{table}[H]
		\centering
		\sisetup{table-format=6.0}
		\setlength\cmidrulewidth{\heavyrulewidth}
		\caption{\small Novel top 5\% DS ($\downarrow$) compared with SOTA models\textsuperscript{\emph{a}}}
		\label{table:novel_ds_selfies}
		\begin{threeparttable}
		\tiny
		\begin{tabular}{lccccc}
			\toprule
			\multirow{2}*{\tabincell{c}{\small Methods\\}} & \multicolumn{5}{c}{Target proteins}\\
			{} & {PARP1} & {FA7} & {5HT1B} & {BRAF} & {JAK2}\\
			\midrule
			\midrule
			{\small REINVENT\cite{reinvent}} & {{\scriptsize -8.702} $\pm$ 0.523} & {{\scriptsize -7.205} $\pm$ 0.264} & {{\scriptsize -8.770} $\pm$ 0.316} & {{\scriptsize -8.392} $\pm$ 0.400} & {{\scriptsize -8.165} $\pm$ 0.277}\\
			{\small MORLD\cite{morld}} & {{\scriptsize -7.532} $\pm$ 0.260} & {{\scriptsize -6.263} $\pm$ 0.165} & {{\scriptsize -7.869} $\pm$ 0.650} & {{\scriptsize -8.040} $\pm$ 0.337} & {{\scriptsize -7.816} $\pm$ 0.133}\\
			{\small HierVAE\cite{hiervae}} & {{\scriptsize -9.487} $\pm$ 0.278} & {{\scriptsize -6.812} $\pm$ 0.274} & {{\scriptsize -8.081} $\pm$ 0.252} & {{\scriptsize -8.978} $\pm$ 0.525} & {{\scriptsize -8.285} $\pm$ 0.370}\\
			{\small FREED\cite{freed}} & {{\scriptsize -10.427} $\pm$ 0.177} & {{\scriptsize -8.297} $\pm$ 0.094} & {{\scriptsize -10.425} $\pm$ 0.331} & {{\scriptsize -10.325} $\pm$ 0.164} & {{\scriptsize -9.624} $\pm$ 0.102}\\
			{\small GDSS\cite{gdss}} & {{\scriptsize -9.967} $\pm$ 0.028} & {{\scriptsize -7.775} $\pm$ 0.039} & {{\scriptsize -9.459} $\pm$ 0.101} & {{\scriptsize -9.224} $\pm$ 0.068} & {{\scriptsize -8.926} $\pm$ 0.089}\\
			{\small MOOD\cite{mood}} & {{\scriptsize -10.865} $\pm$ 0.113} & {{\scriptsize -8.160} $\pm$ 0.071} & {{\scriptsize -11.145} $\pm$ 0.042} & {{\scriptsize -11.063} $\pm$ 0.034} & {{\scriptsize -10.147} $\pm$ 0.060}\\
			\midrule
			{\small ChemBFN {\scriptsize strategy 3}} & {{\scriptsize\textbf{-12.455}} $\pm$ 0.068} & {{\scriptsize\textbf{-9.527}} $\pm$ 0.033} & {{\scriptsize\textbf{-12.609}} $\pm$ 0.045} & {{\scriptsize\underline{-12.043}} $\pm$ 0.117} & {{\scriptsize\textbf{-11.690}} $\pm$ 0.105}\\
			{\small ChemBFN {\scriptsize strategy 4}} & {{\scriptsize\underline{-12.358}} $\pm$ 0.102} & {{\scriptsize\underline{-9.315}} $\pm$ 0.026} & {{\scriptsize\underline{-12.359}} $\pm$ 0.172} & {{\scriptsize\textbf{-12.061}} $\pm$ 0.087} & {{\scriptsize\underline{-11.666}} $\pm$ 0.089}\\
			\bottomrule
		\end{tabular}
		\begin{tablenotes}
			\small
			\item [a] $\downarrow$ stands for the lower the better while other settings are the same as \prettyref{table:novel_hit_ratio_selfies}
		\end{tablenotes}
		\end{threeparttable}
	\end{table}
	
	\par
\begin{figure}[H]
		\centering
		\tiny
		\begin{tabular}{lw{c}{0.15\linewidth}w{c}{0.15\linewidth}w{c}{0.15\linewidth}w{c}{0.15\linewidth}w{c}{0.15\linewidth}}
			{} & {PARP1} & {FA7} & {5HT1B} & {BRAF} & {JAK2}\\
			{\begin{turn}{90}ZINC250k\end{turn}} & {\includegraphics*[width=0.1\linewidth]{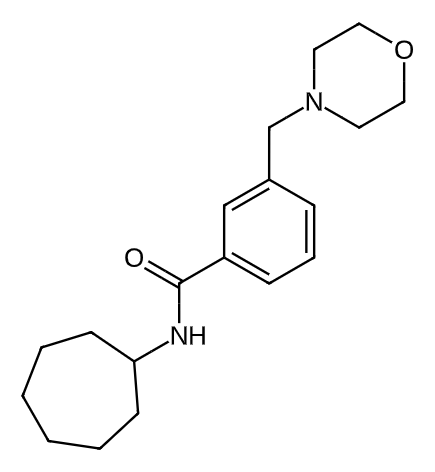}} & {\includegraphics*[width=0.1\linewidth]{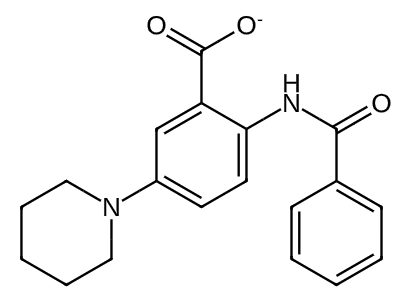}} & {\includegraphics*[width=0.15\linewidth]{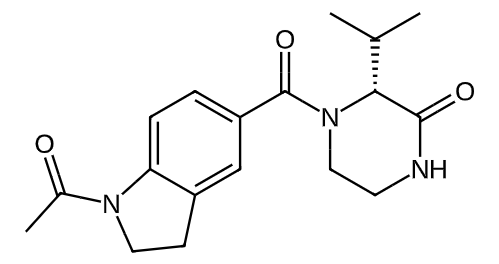}} & {\includegraphics*[width=0.06\linewidth]{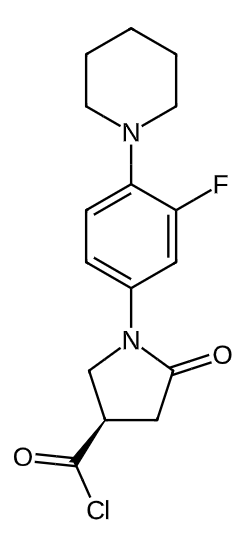}} & {\includegraphics*[width=0.07\linewidth]{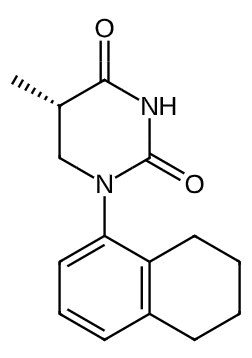}}\\
			{} & {DS = -8.5} & {DS = -7.1} & {DS = -8.9} & {DS = -9.0} & {DS = -8.8}\\
			{\begin{turn}{90}ChemBFN\end{turn}} & {\includegraphics*[width=0.1\linewidth]{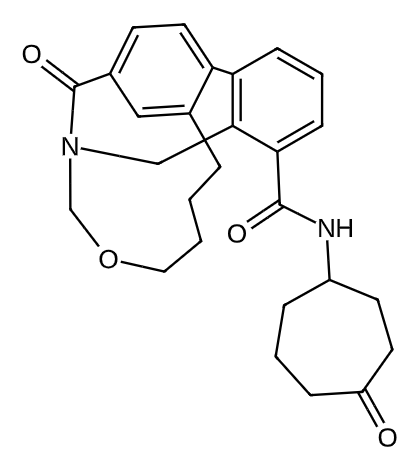}} & {\includegraphics*[width=0.2\linewidth]{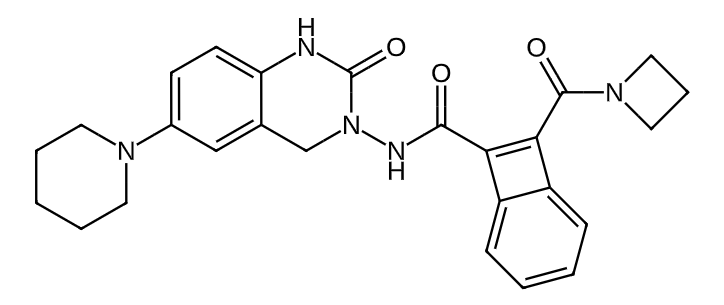}} & {\includegraphics*[width=0.18\linewidth]{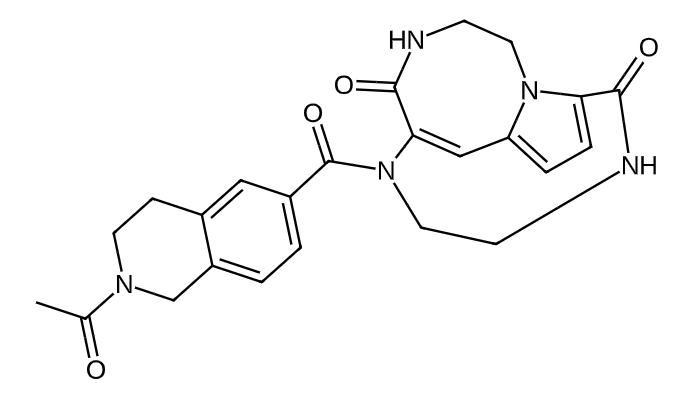}} & {\includegraphics*[width=0.15\linewidth]{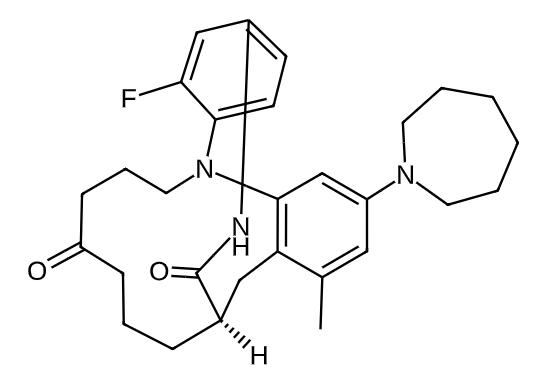}} & {\includegraphics*[width=0.12\linewidth]{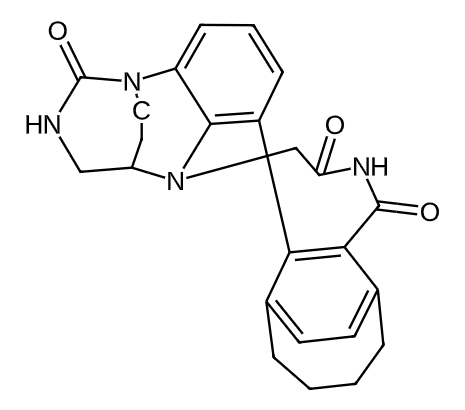}}\\
			{} & \multirow{2}*{\tabincell{c}{DS = -14.9\\similarity = 0.333}} & \multirow{2}*{\tabincell{c}{DS = -10.1\\similarity = 0.353}} & \multirow{2}*{\tabincell{c}{DS = -13.9\\similarity = 0.373}} & \multirow{2}*{\tabincell{c}{DS = -14.4\\similarity = 0.316}} & \multirow{2}*{\tabincell{c}{DS = -13.9\\similarity = 0.318}}\\
			{} & {} & {} & {} & {} & {}\\
		\end{tabular}
		\caption{\small Examples of novel hit molecules generated by ChemBFN (\textbf{SMILES} version) against their closest neighbours in the training space.}
		\label{fig:hit_molecule_smiles}
	\end{figure}

\begin{figure}[H]
		\centering
		\tiny
		\begin{tabular}{lw{c}{0.15\linewidth}w{c}{0.15\linewidth}w{c}{0.15\linewidth}w{c}{0.15\linewidth}w{c}{0.15\linewidth}}
			{} & {PARP1} & {FA7} & {5HT1B} & {BRAF} & {JAK2}\\
			{\begin{turn}{90}ZINC250k\end{turn}} & {\includegraphics*[width=0.08\linewidth]{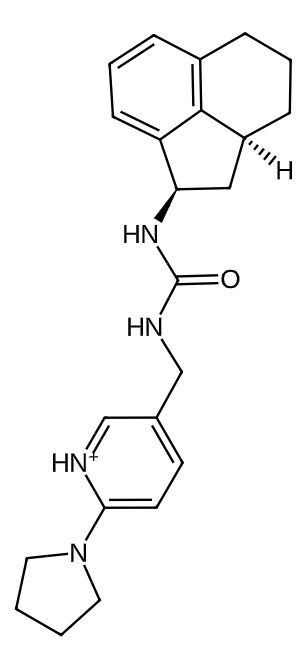}} & {\includegraphics*[width=0.15\linewidth]{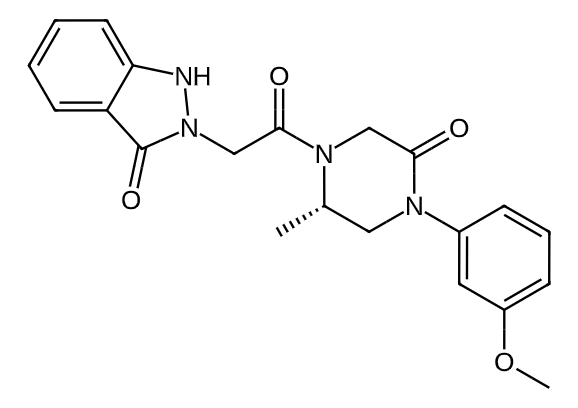}} & {\includegraphics*[width=0.1\linewidth]{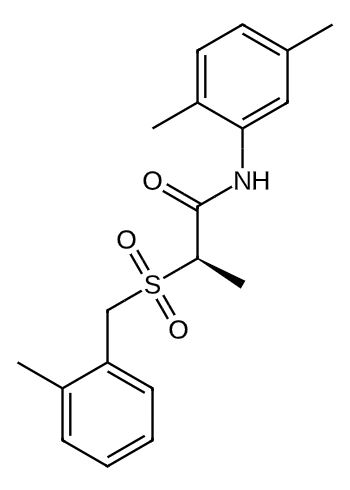}} & {\includegraphics*[width=0.08\linewidth]{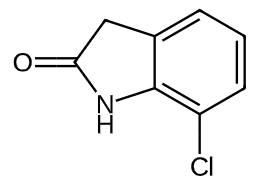}} & {\includegraphics*[width=0.1\linewidth]{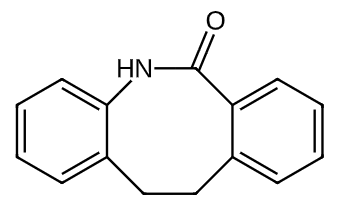}}\\
			{} & {DS = -11.2} & {DS = -7.5} & {DS = -9.6} & {DS = -6.2} & {DS = -7.8}\\
			{\begin{turn}{90}ChemBFN\end{turn}} & {\includegraphics*[width=0.22\linewidth]{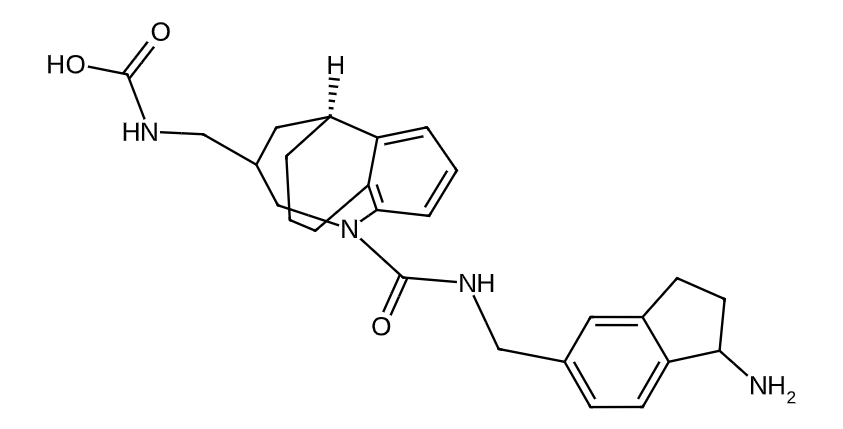}} & {\includegraphics*[width=0.2\linewidth]{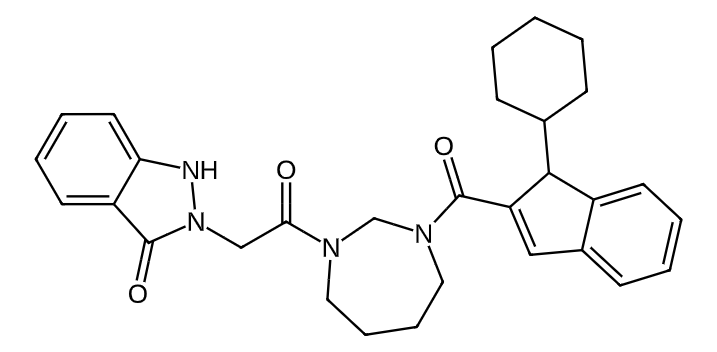}} & {\includegraphics*[width=0.19\linewidth]{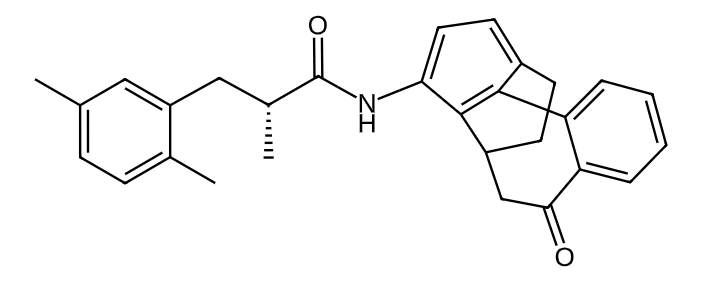}} & {\includegraphics*[width=0.12\linewidth]{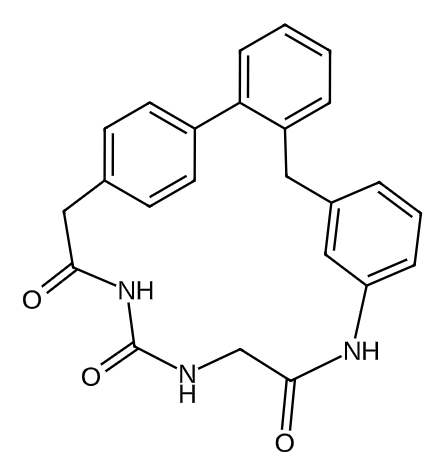}} & {\includegraphics*[width=0.12\linewidth]{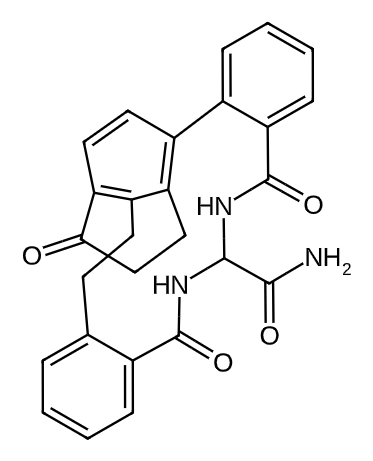}}\\
			{} & \multirow{2}*{\tabincell{c}{DS = -14.0\\similarity = 0.384}} & \multirow{2}*{\tabincell{c}{DS = -10.6\\similarity = 0.357}} & \multirow{2}*{\tabincell{c}{DS = -14.5\\similarity = 0.380}} & \multirow{2}*{\tabincell{c}{DS = -14.0\\similarity = 0.308}} & \multirow{2}*{\tabincell{c}{DS = -13.5\\similarity = 0.383}}\\
			{} & {} & {} & {} & {} & {}\\
		\end{tabular}
		\caption{\small Examples of novel hit molecules generated by ChemBFN (\textbf{SELFIES} version) against their closest neighbours in the training space.}
		\label{fig:hit_molecule_selfies}
	\end{figure}

\begin{figure}[H]
		\centering
		\includegraphics[width=0.8\linewidth]{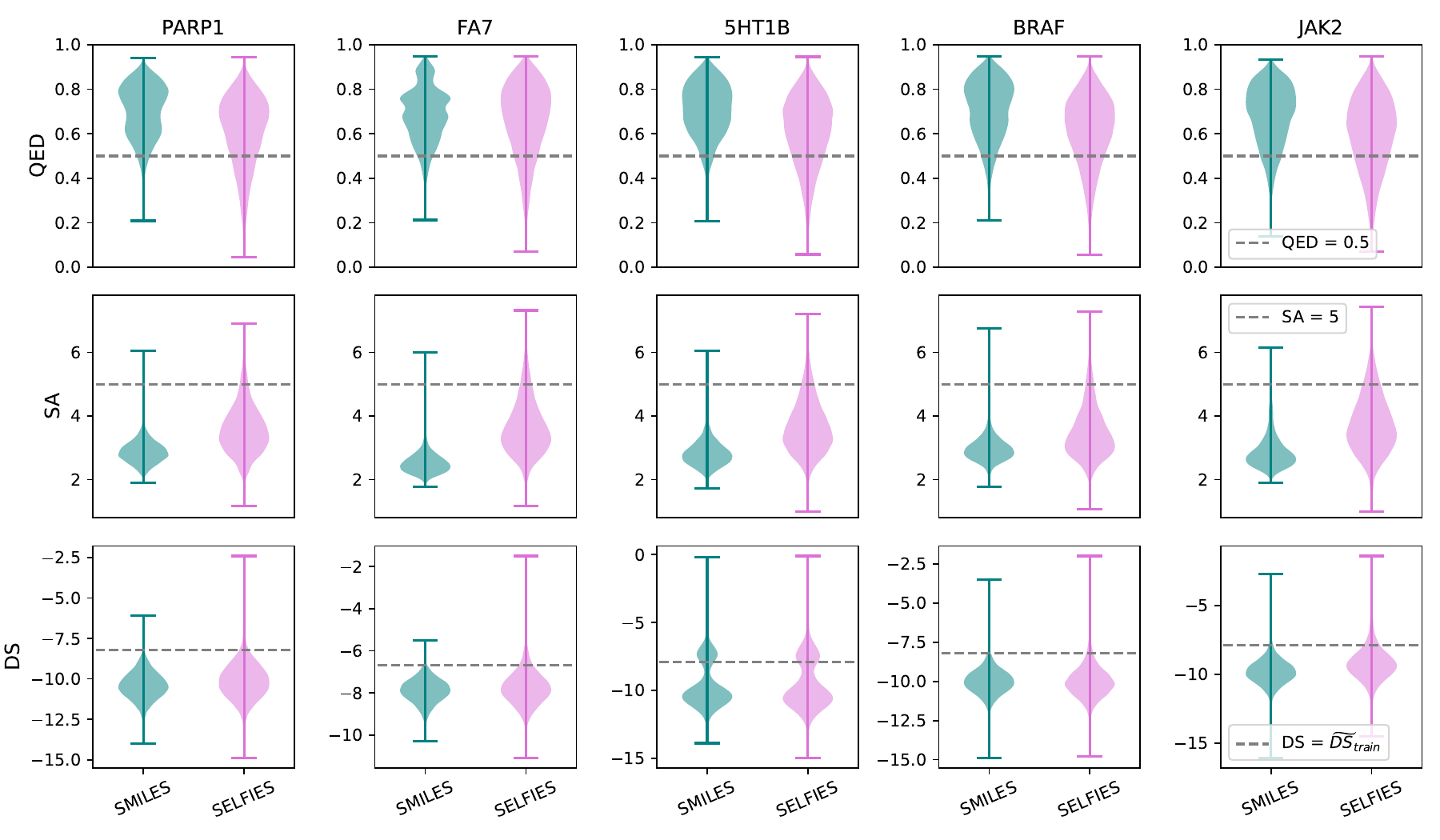}
		\caption{\small The distributions of QED, SA, and DS scores of generated molecules for different target proteins grouped in SMILES or SELFIES.}
		\label{fig:zinc250k_score}
	\end{figure}

 Examples of novel hit molecules generated by our model (both SMILES version and SELFIES version) were illustrated in \prettyref{fig:hit_molecule_smiles} and \prettyref{fig:hit_molecule_selfies}. An interesting observation is that although the ring systems in the training data are generally small, our model tended to generate larger ring systems or even macrocyclic systems that showed lower binding energies. We also reported in \prettyref{table:solve_rate} the total ratio of generated molecules whose retrosynthesis could be solved by AiZynthFinder\cite{aizynthfinder} with default settings. Although an illustration of how the DS, SA, and QED scores of our generated molecules distributed are provided in \prettyref{fig:zinc250k_score}, the readers are also recommended to refer to the supporting materials to view all the generated samples.	
	
	\begin{table}[H]
		\centering
		\sisetup{table-format=6.0}
		\setlength\cmidrulewidth{\heavyrulewidth}
		\caption{\small The total retrosynthetic path solving rate\textsuperscript{\emph{a}} (\%) of generated molecules estimated via AiZynthFinder}
		\label{table:solve_rate}
		\begin{threeparttable}
			\tiny
			\begin{tabular}{lccccc}
				\toprule
				\multirow{2}*{\tabincell{c}{\small Versions\\}} & \multicolumn{5}{c}{Target proteins}\\
				{} & {PARP1} & {FA7} & {5HT1B} & {BRAF} & {JAK2}\\
				\midrule
				\midrule
				{\small SMILES} & {\scriptsize 42.275} & {\scriptsize 61.528} & {\scriptsize 32.625} & {\scriptsize 40.623} & {\scriptsize 21.473}\\
				{\small SELFIES} & {\scriptsize 15.277} & {\scriptsize 15.463} & {\scriptsize 11.746} & {\scriptsize 18.173} & {\scriptsize 11.564}\\
				\bottomrule
			\end{tabular}
			\begin{tablenotes}
				\small
				\item [a] Non-novel, low QED-scored, and high SA-scored molecules were not filtered out when calculating the ratios.
			\end{tablenotes}
		\end{threeparttable}
	\end{table}
	
	\subsection{Conditional Generation of Protein Sequences}
	\par
\begin{figure}[H]
		\centering
		\includegraphics[width=0.9\linewidth]{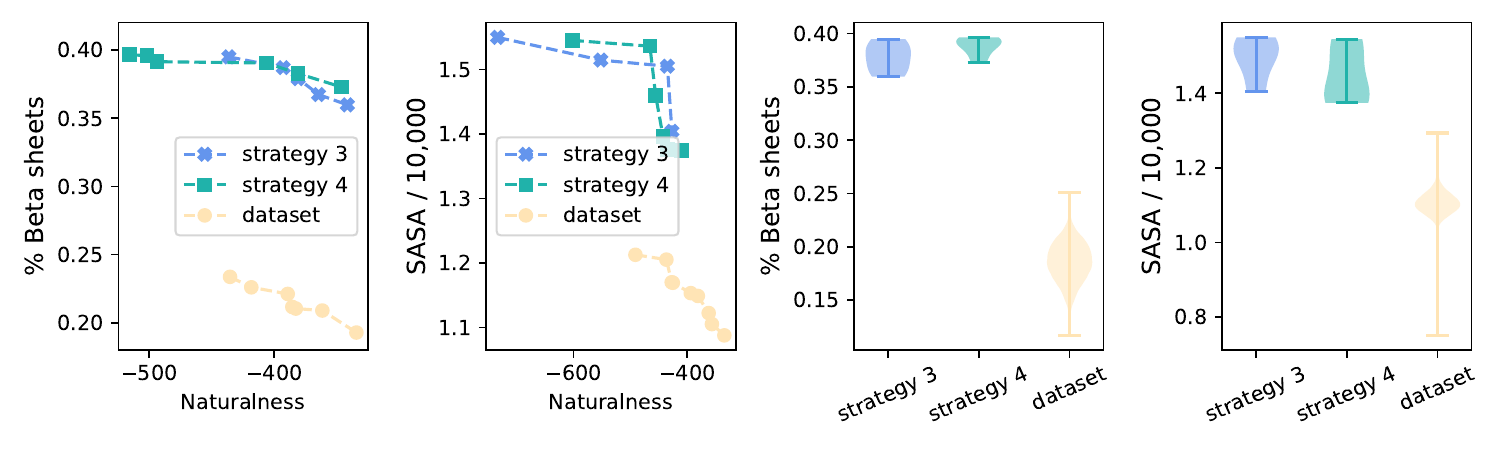}
		\caption{\small (Left 2 plots) The Pareto fronts of objective values \textit{v.s.} naturalness of generated protein sequences and dataset; (Right 2 plots) The violin plots of the objective values of generated samples and dataset. Notice that our method generated proteins with higher objective values than any entity in the training dataset while maintaining the similar naturalness to natural proteins.}
		\label{fig:protein_result}
	\end{figure}

 An amino acid-wise tokenizer was added to the original ChemBFN model to enable protein sequence generating. We trained one model to optimise the percentage of beta sheets and one model to optimise SASA. Each model, after training, generated $32\times 2$ protein sequences (half was generated via strategy 3 and half via strategy 4) guided by an objective value pointing to the higher property regions. As shown in \prettyref{fig:protein_result}, the generated samples all had higher objective values than the training space. When estimated the naturalness (ProtGPT2\cite{protgpt2} log likelihood as suggested by L. Klarner \textit{et al}\cite{cgd}) of the generated proteins, we found that models utilising either strategy 3 or strategy 4 gave reasonably acceptable results compared with natural proteins (see \prettyref{fig:protein_result}). As SASA and the percentage of beta sheets of proteins are highly correlated their structures, we conclude that our model is capable of determining the relationship between an objective (scalar) value to its corresponding chemical structures, unsupervised, and extrapolating to unseen spaces. However, the readers should also notice that when the properties went farer away from the training data, the validities had some decrease, i.e., the naturalness went more negative (\prettyref{fig:protein_result} left 2 plots).
	
	\subsection{What If You Care More About In-Distribution Sampling}
	\par
\begin{figure}[H]
		\centering
		\includegraphics[width=0.9\linewidth]{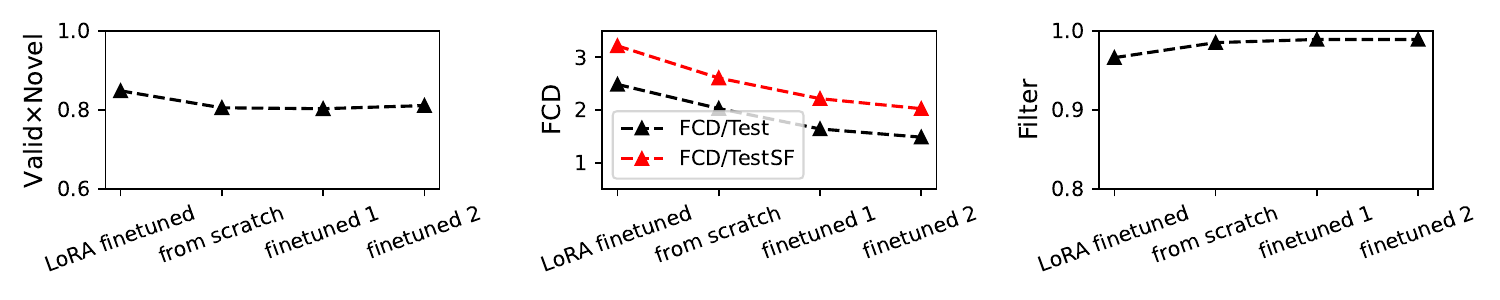}
		\caption{\small The key MOSES testing metrics of finetuned models. We reported $\mathrm{Valid}\times\mathrm{Novel}$ values instead of validity and novelty separately. Note that we employed strategy 1 in this experiment.}
		\label{fig:moses_metrics_ft}
	\end{figure}

 In this section, we demonstrate how pre-training affects the sampling distributions of MOSES as an example. Finetuned models based on models, provided by N. Tao \textit{et al}\cite{chembfn}, pretrained on 40M and 190M molecules selected from ZINC15 database\cite{zinc15} were denoted as `finetuned 1' and `finetuned 2', respectively. As shown in \prettyref{fig:moses_metrics_ft}, the increasing of pre-training data surprisingly led to the decrease of FCD metrics. However, when applying LoRA\cite{lora} finetune strategy (rank = 4, $\alpha=1$) to the pretrainind model (pretrained on 190M molecules), both FCD and the ratio of novel generated molecules increased while the ratio of molecules which passed the MOSES benchmark filters decreased. We conclude here that larger scale pre-training helps to obtain generated molecules closer to the training space only when all the parameters of the network were finetuned; on the other hand, LoRA finetuning, i.e., reduced parameter finetuning, will further strengthen the OOD-ness.
	
	\subsection{Computational Details}
	\par All the models were trained on single Nvidia A100 GPU with a batch-size of 120 for MOSES and GuacaMol datasets and 128 for ZINC250k and protein datasets for 100 epochs. The learning rate was $5\times 10^{-5}$ which was linearly increased from $10^{-8}$ during the first 1,000 steps. AdamW\cite{adamw} method with default parameters was employed to optimise the model weights. An unconditional rate of 0.2 was chosen when training conditional models. During sampling process, the batch-size was 3,000 for small molecules and 32 for protein sequences; the guidance strength was 0.5 for small molecules and 1.0 for protein sequences when a guidance vector was applied. The number of sampling steps, if not specified, was 100 for MOSES while 1,000 for GuacaMol, ZINC250k and protein sequences. The models trained for generating small molecules had 12 layers, 8 attention heads per layer and 512 hidden feature sizes (54.5M of total parameters); the models for protein sequences had 12 layers, 16 attention heads per layer and 1,024 hidden feature sizes (216M of total parameters).
	\par RDKit\cite{rdkit} was used to generate 3-dimensional conformations and Morgan fingerprint (radius = 2, dimension = 1024) from SMILES strings and calculate the Tanimoto similarity, QED, and SA quantities. The DS were computed via QuickVina 2\cite{qvina2}. The properties of proteins were calculated via Biopython\cite{biopython} package version 1.84; to obtain the solvent accessible surface area, the 3-dimensional structures were predicted by ESMFold\cite{esmfold} model first, which was followed by Shrake-Rupley algorithm\cite{shrake-and-rupley} calculations.
	\par In order to generate UMAP plots, the vector molecular representations were extracted from the last activation layer of ChemNet\cite{fcd}, which were later projected to 2-dimensional vectors by UMAP package\cite{umap}. 5,000 random molecules were selected from the dataset as the representatives.
	
	\section{Theoretical Analysis of OOD Generation in ChemBFN with SAR}
	\par Pointed out by M. Kamb \text{et al}\cite{creativity}, the translational equivariance \textit{w.r.t} the pixel space and the locality characteristics of diffusion-style models are two, but not all, important aspects that result to the creativity of image generation models, i.e., generating images out side the distribution of training data. In the context of molecular generation, we have an equivalent property of permutational equivariance of atomic orders when attention-based models are employed because of the permutational equivariant characteristic of attention mechanism\cite{attention} (if we ignore the positional embedding). However, the global attention mechanism lacks the built-in property of locality, obviously, because of its non-local nature.
	\par
\begin{figure}[H]
		\centering
		\includegraphics[width=0.9\linewidth]{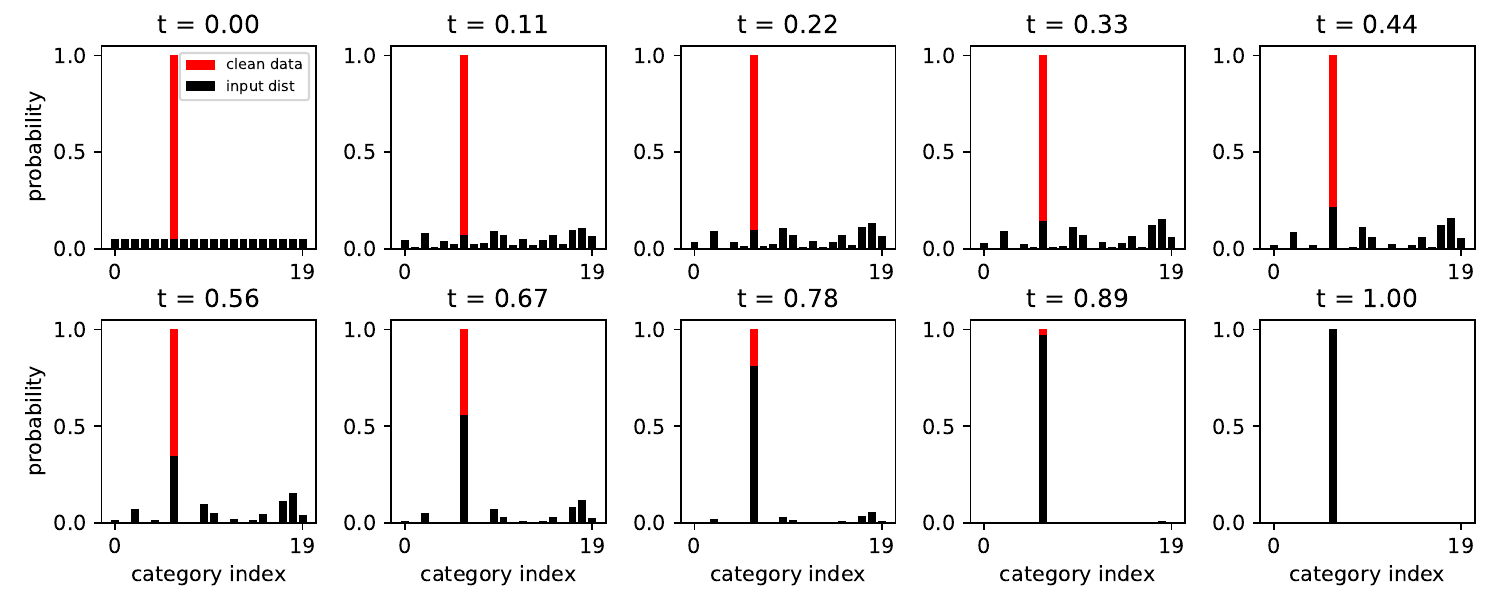}
		\caption{\small An example of the input distributions at different time $t$ with K = 20. Notice that the distribution developed from an uniform distribution at $t = 0$ to a sharp line at $t = 1$.}
		\label{fig:input_dist}
	\end{figure}

\begin{figure}[H]
		\centering
		\includegraphics[width=0.8\linewidth]{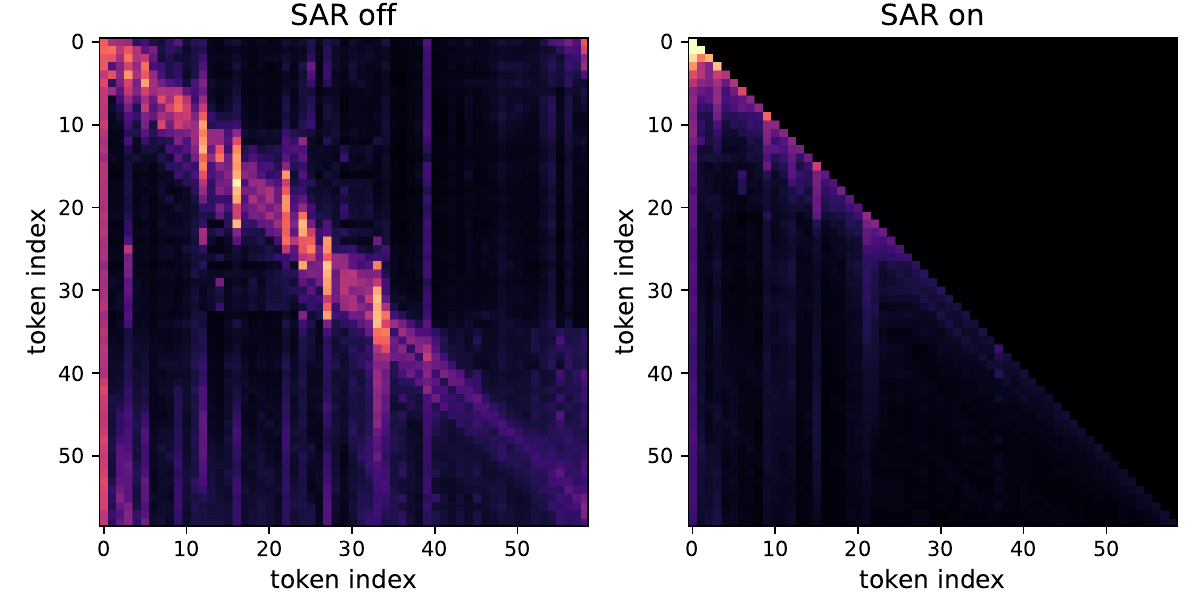}
		\caption{\small A visualisation of the attention maps (averaged over all layers of all attention heads) when SAR mechanism, i.e., the casual masking, is switched on and off during inference. The brighter the larger attention scores. Notice that the large-value regions are all close to the main diagonal.}
		\label{fig:attention_map}
	\end{figure}

 In the framework of BFN, the locality can be achieved as a learnt property despite the non-locality of attention mechanism. Considering the training objective of discrete BFN\cite{bfn,chembfn}
	\begin{equation}
		L^{\infty}(\boldsymbol{x}) = \frac{K}{2}\mathbb{E}_{t\sim U(0,1)}\left(\alpha(t)\|\boldsymbol{e_{x}}-\hat{\boldsymbol{e}(\boldsymbol{\theta};t)}\|^{2}\right),
	\label{eq:bfn_loss}
	\end{equation}
	where K is the number of categories, $\alpha(t) = d\beta(t)/dt$ is a monotonic increasing function (therefore $\beta(t)$ is also monotonic increasing), $\boldsymbol{e_{x}}$ is the categorical
	distribution of clean data $\boldsymbol{x}$, and $\hat{\boldsymbol{e}(\boldsymbol{\theta};t)}$ is computed as the output of the neural network that toke an input distribution $\boldsymbol{p}_{I}(\boldsymbol{x}|\boldsymbol{\theta}) = softmax\left(\beta(t)(K\boldsymbol{e_{x}}-1)+\sqrt{K\beta(t)}\boldsymbol{\epsilon}\right)_{dim=-1}$ ($\boldsymbol{\epsilon}\sim \mathcal{N}(\boldsymbol{0},\boldsymbol{I})$; see \prettyref{fig:chembfn_flowchart}) as the input at time $t$. When $t$ approaches to 1, the entropy of the input distribution becomes small, i.e., the peak becomes sharp, and almost identical to the clean data (see \prettyref{fig:input_dist}). Hence at these regions, to minimise \prettyref{eq:bfn_loss} the neural network only need to fit an identical mapping (or anything close to identical), which leads to attention scores concentrate in the regions close to the main diagonals of the attention maps. Since the non-linear $\alpha(t)$ used in ChemBFN model was designed to increase much faster when $t$ is close to 1 than other intervals\cite{chembfn}, the force pushing the model to achieve locality further outweighs the non-local nature. When SAR is enabled, the locality is further enhanced as half of the non-local relationships are removed by the causal mask. In \prettyref{fig:attention_map}, we showed that the observations consisted with our theory.
	
	
	\par On the other hand, masked diffusion language models (MDLM) that can generate text-based molecular structures use cross entropy as the main component of their training objectives\cite{md4,llada,simple_mdlm,radd}. Although different MDLM use slightly different formulae, for the sake of simplicity, we will use the training objective of LLaDA model\cite{llada} as an example: for a model that takes $\boldsymbol{x}_{t} = (^{0}\boldsymbol{x}_{t},^{1}\boldsymbol{x}_{t},^{2}\boldsymbol{x}_{t},...)$ ($t\in(0,1]$) in which $^{k}\boldsymbol{x}_{t}$ has probability $t$ of being masked and probability $1-t$ of remaining unmasked and outputs $\boldsymbol{p}(\cdot | \boldsymbol{x}_{t})$ as the predicted probabilities of the masked tokens, the loss function is defined as
	\begin{equation}
		L^{\infty}(\boldsymbol{x}_{t},\boldsymbol{x}_{0}) = -\mathbb{E}_{t\sim U(0,1)}\left(\frac{1}{t}\sum_{k}^{^{k}\boldsymbol{x}_{t}~\rm is~masked}\log(^{k}\boldsymbol{p}(\boldsymbol{x}_{0}|\boldsymbol{x}_{t}))\right),
	\end{equation}
	where $\boldsymbol{x}_{0}$ is the clean data. The readers may notice that only the masked tokens are used to calculate the loss, thus the model is not encouraged to fit an identical mapping at any point, which less guarantees the locality. To empirically show the differences possessed by the different mathematics, we modified the input layer to make it compatible with LLaDA training objective and trained the modified model on the ZINC250k dataset utilising the identical settings when training ChemBFN models. The FCD scores between the learning data and the unconditional generated molecules following strategy 1, strategy 2, strategy 3, and strategy 4 were 3.99, 4.42, 3.62, and 4.10, respectively. The difference of the distributional distances possessed by switching on and off SAR was as small as 9.7\%, compared with a difference of 44.5\% in the case of ChemBFN, which proved that the enforced learnt locality is the key difference between a BFN model and a diffusion model, while SAR helps OOD-ness \textit{iff} the locality is well present.
	\par The mathematics of BFN and the new accuracy schedule in ChemBFN guarantee the locality of the model regardless of the global attention, which enables the model to segment molecules into smaller substructures than that obtained by diffusion models, to precisely learn the relationships between the molecular properties and those substructures during training, and to combine them into novel molecules during inference. We believe this characteristic brings ChemBFN model desired OOD performance.
	
	\section{Conclusion}
	\par In this research, we showed that BFN, especially ChemBFN model, is naturally a controllable OOD sampler, which is versatile to generate both small molecules and large chemical systems such as proteins. We found that for unconditional generation, the normal training-SAR sampling strategy promoted the OOD-ness of the model most; when a guidance was switched on, the OOD behaviour was pushed to a higher level, in which case the SAR trained models outperformed several state-of-the-art methods: the SAR training-normal sampling strategy tended to samples with higher objective values while the SAR training-SAR sampling strategy had a tendency to yield more novel samples. The generated samples in the OOD regions satisfied several real-world requirements including high drug-likeness, low synthetic difficulties, and similar naturalness to their natural counterparts, which makes our method a strong candidate to handle \textit{de novo} drug design challenges. In addition, we developed strategies to reduce the inferencing time which makes ChemBFN more cost efficient in large scale sampling.
	
	\section{Data and Software Availability}
	\par The code of ChemBFN and instructions necessary to reproduce the results of this study are available for downloading at:
	\newline
	\url{https://github.com/Augus1999/bayesian-flow-network-for-chemistry}.
	\par The pre-trained models can be accessed at: \url{https://huggingface.co/suenoomozawa/ChemBFN}.
	\par A web-based UI is also available for inferencing: \url{https://github.com/Augus1999/ChemBFN-WebUI}.
	
	\section{Acknowledgements}
	\par The computational source of GPU was provided by Research Center for Computational Science, Okazaki, Japan (Project: 24-IMS-C043).
	
	\section{Conflict of Interest}
	\par The authors claim no conflicts of interest.
	
	\section{Funding Sources}
	\par The authors claim that there is no funding related to this research.
	
	\section{Authors' Contributions}
	\par N.T. and M.A. jointly conducted the project and wrote the manuscript.
	
	\bibliography{chembfn_ood}

\providecommand{\latin}[1]{#1}
\makeatletter
\providecommand{\doi}
  {\begingroup\let\do\@makeother\dospecials
  \catcode`\{=1 \catcode`\}=2 \doi@aux}
\providecommand{\doi@aux}[1]{\endgroup\texttt{#1}}
\makeatother
\providecommand*\mcitethebibliography{\thebibliography}
\csname @ifundefined\endcsname{endmcitethebibliography}
  {\let\endmcitethebibliography\endthebibliography}{}
\begin{mcitethebibliography}{72}
\providecommand*\natexlab[1]{#1}
\providecommand*\mciteSetBstSublistMode[1]{}
\providecommand*\mciteSetBstMaxWidthForm[2]{}
\providecommand*\mciteBstWouldAddEndPuncttrue
  {\def\EndOfBibitem{\unskip.}}
\providecommand*\mciteBstWouldAddEndPunctfalse
  {\let\EndOfBibitem\relax}
\providecommand*\mciteSetBstMidEndSepPunct[3]{}
\providecommand*\mciteSetBstSublistLabelBeginEnd[3]{}
\providecommand*\EndOfBibitem{}
\mciteSetBstSublistMode{f}
\mciteSetBstMaxWidthForm{subitem}{(\alph{mcitesubitemcount})}
\mciteSetBstSublistLabelBeginEnd
  {\mcitemaxwidthsubitemform\space}
  {\relax}
  {\relax}

\bibitem[Bohacek \latin{et~al.}(1996)Bohacek, McMartin, and
  Guida]{drug-design-art}
Bohacek,~R.~S.; McMartin,~C.; Guida,~W.~C. The art and practice of
  structure-based drug design: a molecular modeling perspective.
  \emph{Medicinal research reviews} \textbf{1996}, \emph{16}, 3--50\relax
\mciteBstWouldAddEndPuncttrue
\mciteSetBstMidEndSepPunct{\mcitedefaultmidpunct}
{\mcitedefaultendpunct}{\mcitedefaultseppunct}\relax
\EndOfBibitem
\bibitem[Ertl(2003)]{organic-substituents}
Ertl,~P. Cheminformatics analysis of organic substituents: identification of
  the most common substituents, calculation of substituent properties, and
  automatic identification of drug-like bioisosteric groups. \emph{Journal of
  chemical information and computer sciences} \textbf{2003}, \emph{43},
  374--380\relax
\mciteBstWouldAddEndPuncttrue
\mciteSetBstMidEndSepPunct{\mcitedefaultmidpunct}
{\mcitedefaultendpunct}{\mcitedefaultseppunct}\relax
\EndOfBibitem
\bibitem[Reymond \latin{et~al.}(2010)Reymond, van Deursen, Blum, and
  Ruddigkeit]{chemical-space}
Reymond,~J.-L.; van Deursen,~R.; Blum,~L.~C.; Ruddigkeit,~L. Chemical space as
  a source for new drugs. \emph{Med. Chem. Commun.} \textbf{2010}, \emph{1},
  30--38\relax
\mciteBstWouldAddEndPuncttrue
\mciteSetBstMidEndSepPunct{\mcitedefaultmidpunct}
{\mcitedefaultendpunct}{\mcitedefaultseppunct}\relax
\EndOfBibitem
\bibitem[Abdelraheem \latin{et~al.}(2018)Abdelraheem, Shaabani, and
  D{\"o}mling]{mcr-drug}
Abdelraheem,~E.~M.; Shaabani,~S.; D{\"o}mling,~A. Macrocycles: MCR synthesis
  and applications in drug discovery. \emph{Drug Discovery Today: Technologies}
  \textbf{2018}, \emph{29}, 11--17\relax
\mciteBstWouldAddEndPuncttrue
\mciteSetBstMidEndSepPunct{\mcitedefaultmidpunct}
{\mcitedefaultendpunct}{\mcitedefaultseppunct}\relax
\EndOfBibitem
\bibitem[Garcia~Jimenez \latin{et~al.}(2023)Garcia~Jimenez, Poongavanam, and
  Kihlberg]{macrocycles-in-drug-discovery}
Garcia~Jimenez,~D.; Poongavanam,~V.; Kihlberg,~J. Macrocycles in Drug
  Discovery—Learning from the Past for the Future. \emph{Journal of Medicinal
  Chemistry} \textbf{2023}, \emph{66}, 5377--5396, PMID: 37017513\relax
\mciteBstWouldAddEndPuncttrue
\mciteSetBstMidEndSepPunct{\mcitedefaultmidpunct}
{\mcitedefaultendpunct}{\mcitedefaultseppunct}\relax
\EndOfBibitem
\bibitem[Driggers \latin{et~al.}(2008)Driggers, Hale, Lee, and
  Terrett]{exploration-of-macrocycles}
Driggers,~E.~M.; Hale,~S.~P.; Lee,~J.; Terrett,~N.~K. The exploration of
  macrocycles for drug discovery—an underexploited structural class.
  \emph{Nature Reviews Drug Discovery} \textbf{2008}, \emph{7}, 608--624\relax
\mciteBstWouldAddEndPuncttrue
\mciteSetBstMidEndSepPunct{\mcitedefaultmidpunct}
{\mcitedefaultendpunct}{\mcitedefaultseppunct}\relax
\EndOfBibitem
\bibitem[Maynard~Smith(1970)]{protein-space}
Maynard~Smith,~J. Natural selection and the concept of a protein space.
  \emph{Nature} \textbf{1970}, \emph{225}, 563--564\relax
\mciteBstWouldAddEndPuncttrue
\mciteSetBstMidEndSepPunct{\mcitedefaultmidpunct}
{\mcitedefaultendpunct}{\mcitedefaultseppunct}\relax
\EndOfBibitem
\bibitem[G{\'o}mez-Bombarelli \latin{et~al.}(2018)G{\'o}mez-Bombarelli, Wei,
  Duvenaud, Hern{\'a}ndez-Lobato, S{\'a}nchez-Lengeling, Sheberla,
  Aguilera-Iparraguirre, Hirzel, Adams, and Aspuru-Guzik]{smiles-vae}
G{\'o}mez-Bombarelli,~R.; Wei,~J.~N.; Duvenaud,~D.;
  Hern{\'a}ndez-Lobato,~J.~M.; S{\'a}nchez-Lengeling,~B.; Sheberla,~D.;
  Aguilera-Iparraguirre,~J.; Hirzel,~T.~D.; Adams,~R.~P.; Aspuru-Guzik,~A.
  Automatic chemical design using a data-driven continuous representation of
  molecules. \emph{ACS central science} \textbf{2018}, \emph{4}, 268--276\relax
\mciteBstWouldAddEndPuncttrue
\mciteSetBstMidEndSepPunct{\mcitedefaultmidpunct}
{\mcitedefaultendpunct}{\mcitedefaultseppunct}\relax
\EndOfBibitem
\bibitem[Lee \latin{et~al.}(2023)Lee, Jo, and Hwang]{mood}
Lee,~S.; Jo,~J.; Hwang,~S.~J. Exploring Chemical Space with Score-based
  Out-of-distribution Generation. 2023;
  \url{https://arxiv.org/abs/2206.07632}\relax
\mciteBstWouldAddEndPuncttrue
\mciteSetBstMidEndSepPunct{\mcitedefaultmidpunct}
{\mcitedefaultendpunct}{\mcitedefaultseppunct}\relax
\EndOfBibitem
\bibitem[Lim \latin{et~al.}(2018)Lim, Ryu, Kim, and Kim]{conditional-vae}
Lim,~J.; Ryu,~S.; Kim,~J.~W.; Kim,~W.~Y. Molecular generative model based on
  conditional variational autoencoder for de novo molecular design.
  \emph{Journal of cheminformatics} \textbf{2018}, \emph{10}, 31\relax
\mciteBstWouldAddEndPuncttrue
\mciteSetBstMidEndSepPunct{\mcitedefaultmidpunct}
{\mcitedefaultendpunct}{\mcitedefaultseppunct}\relax
\EndOfBibitem
\bibitem[Schwalbe-Koda and Gómez-Bombarelli(2020)Schwalbe-Koda, and
  Gómez-Bombarelli]{generative-models-for-automatic-chemical-design}
Schwalbe-Koda,~D.; Gómez-Bombarelli,~R. \emph{Machine Learning Meets Quantum
  Physics}; Springer International Publishing, 2020; p 445–467\relax
\mciteBstWouldAddEndPuncttrue
\mciteSetBstMidEndSepPunct{\mcitedefaultmidpunct}
{\mcitedefaultendpunct}{\mcitedefaultseppunct}\relax
\EndOfBibitem
\bibitem[Zhung \latin{et~al.}(2024)Zhung, Kim, and Kim]{deepicl}
Zhung,~W.; Kim,~H.; Kim,~W.~Y. 3D molecular generative framework for
  interaction-guided drug design. \emph{Nature Communications} \textbf{2024},
  \emph{15}, 2688\relax
\mciteBstWouldAddEndPuncttrue
\mciteSetBstMidEndSepPunct{\mcitedefaultmidpunct}
{\mcitedefaultendpunct}{\mcitedefaultseppunct}\relax
\EndOfBibitem
\bibitem[Walters and Murcko(2020)Walters, and Murcko]{assesse-ai}
Walters,~W.~P.; Murcko,~M. Assessing the impact of generative AI on medicinal
  chemistry. \emph{Nature biotechnology} \textbf{2020}, \emph{38},
  143--145\relax
\mciteBstWouldAddEndPuncttrue
\mciteSetBstMidEndSepPunct{\mcitedefaultmidpunct}
{\mcitedefaultendpunct}{\mcitedefaultseppunct}\relax
\EndOfBibitem
\bibitem[Klarner \latin{et~al.}(2024)Klarner, Rudner, Morris, Deane, and
  Teh]{cgd}
Klarner,~L.; Rudner,~T. G.~J.; Morris,~G.~M.; Deane,~C.~M.; Teh,~Y.~W.
  Context-Guided Diffusion for Out-of-Distribution Molecular and Protein
  Design. 2024; \url{https://arxiv.org/abs/2407.11942}\relax
\mciteBstWouldAddEndPuncttrue
\mciteSetBstMidEndSepPunct{\mcitedefaultmidpunct}
{\mcitedefaultendpunct}{\mcitedefaultseppunct}\relax
\EndOfBibitem
\bibitem[Ho \latin{et~al.}(2020)Ho, Jain, and Abbeel]{ddpm}
Ho,~J.; Jain,~A.; Abbeel,~P. Denoising diffusion probabilistic models.
  \emph{Advances in neural information processing systems} \textbf{2020},
  \emph{33}, 6840--6851\relax
\mciteBstWouldAddEndPuncttrue
\mciteSetBstMidEndSepPunct{\mcitedefaultmidpunct}
{\mcitedefaultendpunct}{\mcitedefaultseppunct}\relax
\EndOfBibitem
\bibitem[Song \latin{et~al.}(2022)Song, Meng, and Ermon]{ddim}
Song,~J.; Meng,~C.; Ermon,~S. Denoising Diffusion Implicit Models. 2022;
  \url{https://arxiv.org/abs/2010.02502}\relax
\mciteBstWouldAddEndPuncttrue
\mciteSetBstMidEndSepPunct{\mcitedefaultmidpunct}
{\mcitedefaultendpunct}{\mcitedefaultseppunct}\relax
\EndOfBibitem
\bibitem[Sohl-Dickstein \latin{et~al.}(2015)Sohl-Dickstein, Weiss,
  Maheswaranathan, and Ganguli]{diffusion_probabilistic}
Sohl-Dickstein,~J.; Weiss,~E.; Maheswaranathan,~N.; Ganguli,~S. Deep
  Unsupervised Learning using Nonequilibrium Thermodynamics. Proceedings of the
  32nd International Conference on Machine Learning. Lille, France, 2015; pp
  2256--2265\relax
\mciteBstWouldAddEndPuncttrue
\mciteSetBstMidEndSepPunct{\mcitedefaultmidpunct}
{\mcitedefaultendpunct}{\mcitedefaultseppunct}\relax
\EndOfBibitem
\bibitem[Song \latin{et~al.}(2021)Song, Sohl-Dickstein, Kingma, Kumar, Ermon,
  and Poole]{score-based}
Song,~Y.; Sohl-Dickstein,~J.; Kingma,~D.~P.; Kumar,~A.; Ermon,~S.; Poole,~B.
  Score-Based Generative Modeling through Stochastic Differential Equations.
  2021; \url{https://arxiv.org/abs/2011.13456}\relax
\mciteBstWouldAddEndPuncttrue
\mciteSetBstMidEndSepPunct{\mcitedefaultmidpunct}
{\mcitedefaultendpunct}{\mcitedefaultseppunct}\relax
\EndOfBibitem
\bibitem[Bortoli \latin{et~al.}(2023)Bortoli, Thornton, Heng, and
  Doucet]{diffusion-schrodinger-bridge}
Bortoli,~V.~D.; Thornton,~J.; Heng,~J.; Doucet,~A. Diffusion Schr\"odinger
  Bridge with Applications to Score-Based Generative Modeling. 2023;
  \url{https://arxiv.org/abs/2106.01357}\relax
\mciteBstWouldAddEndPuncttrue
\mciteSetBstMidEndSepPunct{\mcitedefaultmidpunct}
{\mcitedefaultendpunct}{\mcitedefaultseppunct}\relax
\EndOfBibitem
\bibitem[Frans \latin{et~al.}(2024)Frans, Hafner, Levine, and
  Abbeel]{shortcutmodels}
Frans,~K.; Hafner,~D.; Levine,~S.; Abbeel,~P. One Step Diffusion via Shortcut
  Models. 2024; \url{https://arxiv.org/abs/2410.12557}\relax
\mciteBstWouldAddEndPuncttrue
\mciteSetBstMidEndSepPunct{\mcitedefaultmidpunct}
{\mcitedefaultendpunct}{\mcitedefaultseppunct}\relax
\EndOfBibitem
\bibitem[Lipman \latin{et~al.}(2023)Lipman, Chen, Ben-Hamu, Nickel, and
  Le]{flow-matching}
Lipman,~Y.; Chen,~R. T.~Q.; Ben-Hamu,~H.; Nickel,~M.; Le,~M. Flow Matching for
  Generative Modeling. 2023; \url{https://arxiv.org/abs/2210.02747}\relax
\mciteBstWouldAddEndPuncttrue
\mciteSetBstMidEndSepPunct{\mcitedefaultmidpunct}
{\mcitedefaultendpunct}{\mcitedefaultseppunct}\relax
\EndOfBibitem
\bibitem[Albergo \latin{et~al.}(2023)Albergo, Boffi, and
  Vanden-Eijnden]{flow-and-diffusion}
Albergo,~M.~S.; Boffi,~N.~M.; Vanden-Eijnden,~E. Stochastic Interpolants: A
  Unifying Framework for Flows and Diffusions. 2023;
  \url{https://arxiv.org/abs/2303.08797}\relax
\mciteBstWouldAddEndPuncttrue
\mciteSetBstMidEndSepPunct{\mcitedefaultmidpunct}
{\mcitedefaultendpunct}{\mcitedefaultseppunct}\relax
\EndOfBibitem
\bibitem[Tong \latin{et~al.}(2024)Tong, Fatras, Malkin, Huguet, Zhang,
  Rector-Brooks, Wolf, and Bengio]{ot-cfm}
Tong,~A.; Fatras,~K.; Malkin,~N.; Huguet,~G.; Zhang,~Y.; Rector-Brooks,~J.;
  Wolf,~G.; Bengio,~Y. Improving and generalizing flow-based generative models
  with minibatch optimal transport. 2024;
  \url{https://arxiv.org/abs/2302.00482}\relax
\mciteBstWouldAddEndPuncttrue
\mciteSetBstMidEndSepPunct{\mcitedefaultmidpunct}
{\mcitedefaultendpunct}{\mcitedefaultseppunct}\relax
\EndOfBibitem
\bibitem[Dunn and Koes(2024)Dunn, and Koes]{simplexflow}
Dunn,~I.; Koes,~D.~R. Mixed Continuous and Categorical Flow Matching for 3D De
  Novo Molecule Generation. 2024; \url{https://arxiv.org/abs/2404.19739}\relax
\mciteBstWouldAddEndPuncttrue
\mciteSetBstMidEndSepPunct{\mcitedefaultmidpunct}
{\mcitedefaultendpunct}{\mcitedefaultseppunct}\relax
\EndOfBibitem
\bibitem[Stark \latin{et~al.}(2024)Stark, Jing, Wang, Corso, Berger, Barzilay,
  and Jaakkola]{dirichlet-flow}
Stark,~H.; Jing,~B.; Wang,~C.; Corso,~G.; Berger,~B.; Barzilay,~R.;
  Jaakkola,~T. Dirichlet Flow Matching with Applications to DNA Sequence
  Design. 2024; \url{https://arxiv.org/abs/2402.05841}\relax
\mciteBstWouldAddEndPuncttrue
\mciteSetBstMidEndSepPunct{\mcitedefaultmidpunct}
{\mcitedefaultendpunct}{\mcitedefaultseppunct}\relax
\EndOfBibitem
\bibitem[Davis \latin{et~al.}(2024)Davis, Kessler, Petrache, İsmail~İlkan
  Ceylan, Bronstein, and Bose]{fisher-flow}
Davis,~O.; Kessler,~S.; Petrache,~M.; İsmail~İlkan Ceylan; Bronstein,~M.;
  Bose,~A.~J. Fisher Flow Matching for Generative Modeling over Discrete Data.
  2024; \url{https://arxiv.org/abs/2405.14664}\relax
\mciteBstWouldAddEndPuncttrue
\mciteSetBstMidEndSepPunct{\mcitedefaultmidpunct}
{\mcitedefaultendpunct}{\mcitedefaultseppunct}\relax
\EndOfBibitem
\bibitem[Cheng \latin{et~al.}(2025)Cheng, Li, Peng, and Liu]{sfm}
Cheng,~C.; Li,~J.; Peng,~J.; Liu,~G. Categorical Flow Matching on Statistical
  Manifolds. 2025; \url{https://arxiv.org/abs/2405.16441}\relax
\mciteBstWouldAddEndPuncttrue
\mciteSetBstMidEndSepPunct{\mcitedefaultmidpunct}
{\mcitedefaultendpunct}{\mcitedefaultseppunct}\relax
\EndOfBibitem
\bibitem[Gat \latin{et~al.}(2024)Gat, Remez, Shaul, Kreuk, Chen, Synnaeve, Adi,
  and Lipman]{discrete-flow-matching}
Gat,~I.; Remez,~T.; Shaul,~N.; Kreuk,~F.; Chen,~R. T.~Q.; Synnaeve,~G.;
  Adi,~Y.; Lipman,~Y. Discrete Flow Matching. 2024;
  \url{https://arxiv.org/abs/2407.15595}\relax
\mciteBstWouldAddEndPuncttrue
\mciteSetBstMidEndSepPunct{\mcitedefaultmidpunct}
{\mcitedefaultendpunct}{\mcitedefaultseppunct}\relax
\EndOfBibitem
\bibitem[Dunn and Koes(2024)Dunn, and Koes]{flowmol-ctmc}
Dunn,~I.; Koes,~D.~R. Exploring Discrete Flow Matching for 3D De Novo Molecule
  Generation. 2024; \url{https://arxiv.org/abs/2411.16644}\relax
\mciteBstWouldAddEndPuncttrue
\mciteSetBstMidEndSepPunct{\mcitedefaultmidpunct}
{\mcitedefaultendpunct}{\mcitedefaultseppunct}\relax
\EndOfBibitem
\bibitem[Graves \latin{et~al.}(2024)Graves, Srivastava, Atkinson, and
  Gomez]{bfn}
Graves,~A.; Srivastava,~R.~K.; Atkinson,~T.; Gomez,~F. Bayesian Flow Networks.
  2024; \url{https://arxiv.org/abs/2308.07037}\relax
\mciteBstWouldAddEndPuncttrue
\mciteSetBstMidEndSepPunct{\mcitedefaultmidpunct}
{\mcitedefaultendpunct}{\mcitedefaultseppunct}\relax
\EndOfBibitem
\bibitem[Xue \latin{et~al.}(2024)Xue, Zhou, Nie, Min, Zhang, Zhou, and
  Li]{bfn-sde}
Xue,~K.; Zhou,~Y.; Nie,~S.; Min,~X.; Zhang,~X.; Zhou,~J.; Li,~C. Unifying
  Bayesian Flow Networks and Diffusion Models through Stochastic Differential
  Equations. 2024; \url{https://arxiv.org/abs/2404.15766}\relax
\mciteBstWouldAddEndPuncttrue
\mciteSetBstMidEndSepPunct{\mcitedefaultmidpunct}
{\mcitedefaultendpunct}{\mcitedefaultseppunct}\relax
\EndOfBibitem
\bibitem[Atkinson \latin{et~al.}(2024)Atkinson, Barrett, Cameron, Guloglu,
  Greenig, Robinson, Graves, Copoiu, and Laterre]{protbfn}
Atkinson,~T.; Barrett,~T.~D.; Cameron,~S.; Guloglu,~B.; Greenig,~M.;
  Robinson,~L.; Graves,~A.; Copoiu,~L.; Laterre,~A. Protein Sequence Modelling
  with Bayesian Flow Networks. 2024;
  \url{https://www.biorxiv.org/content/early/2024/09/26/2024.09.24.614734}\relax
\mciteBstWouldAddEndPuncttrue
\mciteSetBstMidEndSepPunct{\mcitedefaultmidpunct}
{\mcitedefaultendpunct}{\mcitedefaultseppunct}\relax
\EndOfBibitem
\bibitem[Qu \latin{et~al.}(2024)Qu, Qiu, Song, Gong, Han, Zheng, Zhou, and
  Ma]{molcraft}
Qu,~Y.; Qiu,~K.; Song,~Y.; Gong,~J.; Han,~J.; Zheng,~M.; Zhou,~H.; Ma,~W.-Y.
  MolCRAFT: Structure-Based Drug Design in Continuous Parameter Space. 2024;
  \url{https://arxiv.org/abs/2404.12141}\relax
\mciteBstWouldAddEndPuncttrue
\mciteSetBstMidEndSepPunct{\mcitedefaultmidpunct}
{\mcitedefaultendpunct}{\mcitedefaultseppunct}\relax
\EndOfBibitem
\bibitem[Song \latin{et~al.}(2024)Song, Gong, Zhou, Zheng, Liu, and Ma]{geobfn}
Song,~Y.; Gong,~J.; Zhou,~H.; Zheng,~M.; Liu,~J.; Ma,~W.-Y. Unified Generative
  Modeling of 3D Molecules with Bayesian Flow Networks. The Twelfth
  International Conference on Learning Representations. 2024\relax
\mciteBstWouldAddEndPuncttrue
\mciteSetBstMidEndSepPunct{\mcitedefaultmidpunct}
{\mcitedefaultendpunct}{\mcitedefaultseppunct}\relax
\EndOfBibitem
\bibitem[Tao and Abe(2025)Tao, and Abe]{chembfn}
Tao,~N.; Abe,~M. Bayesian Flow Network Framework for Chemistry Tasks.
  \emph{Journal of Chemical Information and Modeling} \textbf{2025}, \emph{65},
  1178--1187\relax
\mciteBstWouldAddEndPuncttrue
\mciteSetBstMidEndSepPunct{\mcitedefaultmidpunct}
{\mcitedefaultendpunct}{\mcitedefaultseppunct}\relax
\EndOfBibitem
\bibitem[Vaswani \latin{et~al.}(2023)Vaswani, Shazeer, Parmar, Uszkoreit,
  Jones, Gomez, Kaiser, and Polosukhin]{attention}
Vaswani,~A.; Shazeer,~N.; Parmar,~N.; Uszkoreit,~J.; Jones,~L.; Gomez,~A.~N.;
  Kaiser,~L.; Polosukhin,~I. Attention Is All You Need. 2023;
  \url{https://arxiv.org/abs/1706.03762}\relax
\mciteBstWouldAddEndPuncttrue
\mciteSetBstMidEndSepPunct{\mcitedefaultmidpunct}
{\mcitedefaultendpunct}{\mcitedefaultseppunct}\relax
\EndOfBibitem
\bibitem[Weininger(1988)]{smiles}
Weininger,~D. SMILES, a chemical language and information system. 1.
  Introduction to methodology and encoding rules. \emph{Journal of chemical
  information and computer sciences} \textbf{1988}, \emph{28}, 31--36\relax
\mciteBstWouldAddEndPuncttrue
\mciteSetBstMidEndSepPunct{\mcitedefaultmidpunct}
{\mcitedefaultendpunct}{\mcitedefaultseppunct}\relax
\EndOfBibitem
\bibitem[Krenn \latin{et~al.}(2020)Krenn, H{\"a}se, Nigam, Friederich, and
  Aspuru-Guzik]{selfies}
Krenn,~M.; H{\"a}se,~F.; Nigam,~A.; Friederich,~P.; Aspuru-Guzik,~A.
  Self-referencing embedded strings (SELFIES): A 100\% robust molecular string
  representation. \emph{Machine Learning: Science and Technology}
  \textbf{2020}, \emph{1}, 045024\relax
\mciteBstWouldAddEndPuncttrue
\mciteSetBstMidEndSepPunct{\mcitedefaultmidpunct}
{\mcitedefaultendpunct}{\mcitedefaultseppunct}\relax
\EndOfBibitem
\bibitem[Williams(1992)]{reinforce}
Williams,~R.~J. Simple statistical gradient-following algorithms for
  connectionist reinforcement learning. \emph{Machine learning} \textbf{1992},
  \emph{8}, 229--256\relax
\mciteBstWouldAddEndPuncttrue
\mciteSetBstMidEndSepPunct{\mcitedefaultmidpunct}
{\mcitedefaultendpunct}{\mcitedefaultseppunct}\relax
\EndOfBibitem
\bibitem[Sutton \latin{et~al.}(1999)Sutton, McAllester, Singh, and
  Mansour]{policy-gradient}
Sutton,~R.~S.; McAllester,~D.; Singh,~S.; Mansour,~Y. Policy gradient methods
  for reinforcement learning with function approximation. \emph{Advances in
  neural information processing systems} \textbf{1999}, \emph{12},
  1057--1063\relax
\mciteBstWouldAddEndPuncttrue
\mciteSetBstMidEndSepPunct{\mcitedefaultmidpunct}
{\mcitedefaultendpunct}{\mcitedefaultseppunct}\relax
\EndOfBibitem
\bibitem[Devlin \latin{et~al.}(2019)Devlin, Chang, Lee, and Toutanova]{bert}
Devlin,~J.; Chang,~M.-W.; Lee,~K.; Toutanova,~K. BERT: Pre-training of Deep
  Bidirectional Transformers for Language Understanding. 2019;
  \url{https://arxiv.org/abs/1810.04805}\relax
\mciteBstWouldAddEndPuncttrue
\mciteSetBstMidEndSepPunct{\mcitedefaultmidpunct}
{\mcitedefaultendpunct}{\mcitedefaultseppunct}\relax
\EndOfBibitem
\bibitem[Yenduri \latin{et~al.}(2023)Yenduri, M, G, Y, Srivastava, Maddikunta,
  G, Jhaveri, B, Wang, Vasilakos, and Gadekallu]{gpt}
Yenduri,~G.; M,~R.; G,~C.~S.; Y,~S.; Srivastava,~G.; Maddikunta,~P. K.~R.;
  G,~D.~R.; Jhaveri,~R.~H.; B,~P.; Wang,~W.; Vasilakos,~A.~V.; Gadekallu,~T.~R.
  Generative Pre-trained Transformer: A Comprehensive Review on Enabling
  Technologies, Potential Applications, Emerging Challenges, and Future
  Directions. 2023; \url{https://arxiv.org/abs/2305.10435}\relax
\mciteBstWouldAddEndPuncttrue
\mciteSetBstMidEndSepPunct{\mcitedefaultmidpunct}
{\mcitedefaultendpunct}{\mcitedefaultseppunct}\relax
\EndOfBibitem
\bibitem[Polykovskiy \latin{et~al.}(2020)Polykovskiy, Zhebrak,
  Sanchez-Lengeling, Golovanov, Tatanov, Belyaev, Kurbanov, Artamonov,
  Aladinskiy, Veselov, Kadurin, Johansson, Chen, Nikolenko, Aspuru-Guzik, and
  Zhavoronkov]{moses}
Polykovskiy,~D. \latin{et~al.}  Molecular Sets (MOSES): A Benchmarking Platform
  for Molecular Generation Models. 2020;
  \url{https://arxiv.org/abs/1811.12823}\relax
\mciteBstWouldAddEndPuncttrue
\mciteSetBstMidEndSepPunct{\mcitedefaultmidpunct}
{\mcitedefaultendpunct}{\mcitedefaultseppunct}\relax
\EndOfBibitem
\bibitem[Brown \latin{et~al.}(2019)Brown, Fiscato, Segler, and
  Vaucher]{guacamol}
Brown,~N.; Fiscato,~M.; Segler,~M.~H.; Vaucher,~A.~C. GuacaMol: benchmarking
  models for de novo molecular design. \emph{Journal of chemical information
  and modeling} \textbf{2019}, \emph{59}, 1096--1108\relax
\mciteBstWouldAddEndPuncttrue
\mciteSetBstMidEndSepPunct{\mcitedefaultmidpunct}
{\mcitedefaultendpunct}{\mcitedefaultseppunct}\relax
\EndOfBibitem
\bibitem[Preuer \latin{et~al.}(2018)Preuer, Renz, Unterthiner, Hochreiter, and
  Klambauer]{fcd}
Preuer,~K.; Renz,~P.; Unterthiner,~T.; Hochreiter,~S.; Klambauer,~G. Fréchet
  ChemNet Distance: A Metric for Generative Models for Molecules in Drug
  Discovery. \emph{Journal of Chemical Information and Modeling} \textbf{2018},
  \emph{58}, 1736--1741, PMID: 30118593\relax
\mciteBstWouldAddEndPuncttrue
\mciteSetBstMidEndSepPunct{\mcitedefaultmidpunct}
{\mcitedefaultendpunct}{\mcitedefaultseppunct}\relax
\EndOfBibitem
\bibitem[Irwin and Shoichet(2005)Irwin, and Shoichet]{zinc}
Irwin,~J.~J.; Shoichet,~B.~K. ZINC- a free database of commercially available
  compounds for virtual screening. \emph{Journal of chemical information and
  modeling} \textbf{2005}, \emph{45}, 177--182\relax
\mciteBstWouldAddEndPuncttrue
\mciteSetBstMidEndSepPunct{\mcitedefaultmidpunct}
{\mcitedefaultendpunct}{\mcitedefaultseppunct}\relax
\EndOfBibitem
\bibitem[Bickerton \latin{et~al.}(2012)Bickerton, Paolini, Besnard, Muresan,
  and Hopkins]{qed}
Bickerton,~G.~R.; Paolini,~G.~V.; Besnard,~J.; Muresan,~S.; Hopkins,~A.~L.
  Quantifying the chemical beauty of drugs. \emph{Nature chemistry}
  \textbf{2012}, \emph{4}, 90--98\relax
\mciteBstWouldAddEndPuncttrue
\mciteSetBstMidEndSepPunct{\mcitedefaultmidpunct}
{\mcitedefaultendpunct}{\mcitedefaultseppunct}\relax
\EndOfBibitem
\bibitem[Ertl and Schuffenhauer(2009)Ertl, and Schuffenhauer]{sa}
Ertl,~P.; Schuffenhauer,~A. Estimation of synthetic accessibility score of
  drug-like molecules based on molecular complexity and fragment contributions.
  \emph{Journal of cheminformatics} \textbf{2009}, \emph{1}, 8\relax
\mciteBstWouldAddEndPuncttrue
\mciteSetBstMidEndSepPunct{\mcitedefaultmidpunct}
{\mcitedefaultendpunct}{\mcitedefaultseppunct}\relax
\EndOfBibitem
\bibitem[Alhossary \latin{et~al.}(2015)Alhossary, Handoko, Mu, and
  Kwoh]{qvina2}
Alhossary,~A.; Handoko,~S.~D.; Mu,~Y.; Kwoh,~C.-K. Fast, accurate, and reliable
  molecular docking with QuickVina 2. \emph{Bioinformatics} \textbf{2015},
  \emph{31}, 2214--2216\relax
\mciteBstWouldAddEndPuncttrue
\mciteSetBstMidEndSepPunct{\mcitedefaultmidpunct}
{\mcitedefaultendpunct}{\mcitedefaultseppunct}\relax
\EndOfBibitem
\bibitem[Gruver \latin{et~al.}(2024)Gruver, Stanton, Frey, Rudner, Hotzel,
  Lafrance-Vanasse, Rajpal, Cho, and Wilson]{nos}
Gruver,~N.; Stanton,~S.; Frey,~N.; Rudner,~T. G.~J.; Hotzel,~I.;
  Lafrance-Vanasse,~J.; Rajpal,~A.; Cho,~K.; Wilson,~A.~G. Protein design with
  guided discrete diffusion. Proceedings of the 37th International Conference
  on Neural Information Processing Systems. Red Hook, NY, USA, 2024\relax
\mciteBstWouldAddEndPuncttrue
\mciteSetBstMidEndSepPunct{\mcitedefaultmidpunct}
{\mcitedefaultendpunct}{\mcitedefaultseppunct}\relax
\EndOfBibitem
\bibitem[McInnes \latin{et~al.}(2020)McInnes, Healy, and Melville]{umap}
McInnes,~L.; Healy,~J.; Melville,~J. UMAP: Uniform Manifold Approximation and
  Projection for Dimension Reduction. 2020;
  \url{https://arxiv.org/abs/1802.03426}\relax
\mciteBstWouldAddEndPuncttrue
\mciteSetBstMidEndSepPunct{\mcitedefaultmidpunct}
{\mcitedefaultendpunct}{\mcitedefaultseppunct}\relax
\EndOfBibitem
\bibitem[Ho and Salimans(2022)Ho, and Salimans]{classifier-free}
Ho,~J.; Salimans,~T. Classifier-Free Diffusion Guidance. 2022;
  \url{https://arxiv.org/abs/2207.12598}\relax
\mciteBstWouldAddEndPuncttrue
\mciteSetBstMidEndSepPunct{\mcitedefaultmidpunct}
{\mcitedefaultendpunct}{\mcitedefaultseppunct}\relax
\EndOfBibitem
\bibitem[Olivecrona \latin{et~al.}(2017)Olivecrona, Blaschke, Engkvist, and
  Chen]{reinvent}
Olivecrona,~M.; Blaschke,~T.; Engkvist,~O.; Chen,~H. Molecular de-novo design
  through deep reinforcement learning. \emph{Journal of cheminformatics}
  \textbf{2017}, \emph{9}, 48\relax
\mciteBstWouldAddEndPuncttrue
\mciteSetBstMidEndSepPunct{\mcitedefaultmidpunct}
{\mcitedefaultendpunct}{\mcitedefaultseppunct}\relax
\EndOfBibitem
\bibitem[Jeon and Kim(2020)Jeon, and Kim]{morld}
Jeon,~W.; Kim,~D. Autonomous molecule generation using reinforcement learning
  and docking to develop potential novel inhibitors. \emph{Scientific reports}
  \textbf{2020}, \emph{10}, 22104\relax
\mciteBstWouldAddEndPuncttrue
\mciteSetBstMidEndSepPunct{\mcitedefaultmidpunct}
{\mcitedefaultendpunct}{\mcitedefaultseppunct}\relax
\EndOfBibitem
\bibitem[Jin \latin{et~al.}(2020)Jin, Barzilay, and Jaakkola]{hiervae}
Jin,~W.; Barzilay,~D.; Jaakkola,~T. Hierarchical Generation of Molecular Graphs
  using Structural Motifs. Proceedings of the 37th International Conference on
  Machine Learning. 2020; pp 4839--4848\relax
\mciteBstWouldAddEndPuncttrue
\mciteSetBstMidEndSepPunct{\mcitedefaultmidpunct}
{\mcitedefaultendpunct}{\mcitedefaultseppunct}\relax
\EndOfBibitem
\bibitem[Yang \latin{et~al.}(2021)Yang, Hwang, Lee, Ryu, and Hwang]{freed}
Yang,~S.; Hwang,~D.; Lee,~S.; Ryu,~S.; Hwang,~S.~J. Hit and Lead Discovery with
  Explorative RL and Fragment-based Molecule Generation. 2021;
  \url{https://arxiv.org/abs/2110.01219}\relax
\mciteBstWouldAddEndPuncttrue
\mciteSetBstMidEndSepPunct{\mcitedefaultmidpunct}
{\mcitedefaultendpunct}{\mcitedefaultseppunct}\relax
\EndOfBibitem
\bibitem[Jo \latin{et~al.}(2022)Jo, Lee, and Hwang]{gdss}
Jo,~J.; Lee,~S.; Hwang,~S.~J. Score-based Generative Modeling of Graphs via the
  System of Stochastic Differential Equations. 2022;
  \url{https://arxiv.org/abs/2202.02514}\relax
\mciteBstWouldAddEndPuncttrue
\mciteSetBstMidEndSepPunct{\mcitedefaultmidpunct}
{\mcitedefaultendpunct}{\mcitedefaultseppunct}\relax
\EndOfBibitem
\bibitem[Genheden \latin{et~al.}(2020)Genheden, Thakkar, Chadimova, Reymond,
  Engkvist, and Bjerrum]{aizynthfinder}
Genheden,~S.; Thakkar,~A.; Chadimova,~V.; Reymond,~J.-L.; Engkvist,~O.;
  Bjerrum,~E.~J. AiZynthFinder: A Fast Robust and Flexible Open-Source Software
  for Retrosynthetic Planning. 2020;
  \url{https://chemrxiv.org/doi/abs/10.26434/chemrxiv.12465371.v1}\relax
\mciteBstWouldAddEndPuncttrue
\mciteSetBstMidEndSepPunct{\mcitedefaultmidpunct}
{\mcitedefaultendpunct}{\mcitedefaultseppunct}\relax
\EndOfBibitem
\bibitem[Ferruz \latin{et~al.}(2022)Ferruz, Schmidt, and H{\"o}cker]{protgpt2}
Ferruz,~N.; Schmidt,~S.; H{\"o}cker,~B. ProtGPT2 is a deep unsupervised
  language model for protein design. \emph{Nature communications}
  \textbf{2022}, \emph{13}, 4348\relax
\mciteBstWouldAddEndPuncttrue
\mciteSetBstMidEndSepPunct{\mcitedefaultmidpunct}
{\mcitedefaultendpunct}{\mcitedefaultseppunct}\relax
\EndOfBibitem
\bibitem[Sterling and Irwin(2015)Sterling, and Irwin]{zinc15}
Sterling,~T.; Irwin,~J.~J. ZINC 15--ligand discovery for everyone.
  \emph{Journal of chemical information and modeling} \textbf{2015}, \emph{55},
  2324--2337\relax
\mciteBstWouldAddEndPuncttrue
\mciteSetBstMidEndSepPunct{\mcitedefaultmidpunct}
{\mcitedefaultendpunct}{\mcitedefaultseppunct}\relax
\EndOfBibitem
\bibitem[Hu \latin{et~al.}(2022)Hu, Shen, Wallis, Allen-Zhu, Li, Wang, Wang,
  and Chen]{lora}
Hu,~E.~J.; Shen,~Y.; Wallis,~P.; Allen-Zhu,~Z.; Li,~Y.; Wang,~S.; Wang,~L.;
  Chen,~W. Lo{RA}: Low-Rank Adaptation of Large Language Models. International
  Conference on Learning Representations. 2022\relax
\mciteBstWouldAddEndPuncttrue
\mciteSetBstMidEndSepPunct{\mcitedefaultmidpunct}
{\mcitedefaultendpunct}{\mcitedefaultseppunct}\relax
\EndOfBibitem
\bibitem[Loshchilov and Hutter(2019)Loshchilov, and Hutter]{adamw}
Loshchilov,~I.; Hutter,~F. Fixing Weight Decay Regularization in Adam. 2019;
  \url{https://arxiv.org/abs/1711.05101}\relax
\mciteBstWouldAddEndPuncttrue
\mciteSetBstMidEndSepPunct{\mcitedefaultmidpunct}
{\mcitedefaultendpunct}{\mcitedefaultseppunct}\relax
\EndOfBibitem
\bibitem[rdk()]{rdkit}
RDKit: Open-source cheminformatics. \url{https://www.rdkit.org}, Accessed:
  2024-11-12\relax
\mciteBstWouldAddEndPuncttrue
\mciteSetBstMidEndSepPunct{\mcitedefaultmidpunct}
{\mcitedefaultendpunct}{\mcitedefaultseppunct}\relax
\EndOfBibitem
\bibitem[Cock \latin{et~al.}(2009)Cock, Antao, Chang, Chapman, Cox, Dalke,
  Friedberg, Hamelryck, Kauff, Wilczynski, and de~Hoon]{biopython}
Cock,~P. J.~A.; Antao,~T.; Chang,~J.~T.; Chapman,~B.~A.; Cox,~C.~J.; Dalke,~A.;
  Friedberg,~I.; Hamelryck,~T.; Kauff,~F.; Wilczynski,~B.; de~Hoon,~M. J.~L.
  Biopython: freely available Python tools for computational molecular biology
  and bioinformatics. \emph{Bioinformatics} \textbf{2009}, \emph{25},
  1422--1423\relax
\mciteBstWouldAddEndPuncttrue
\mciteSetBstMidEndSepPunct{\mcitedefaultmidpunct}
{\mcitedefaultendpunct}{\mcitedefaultseppunct}\relax
\EndOfBibitem
\bibitem[Lin \latin{et~al.}(2022)Lin, Akin, Rao, Hie, Zhu, Lu, Smetanin, dos
  Santos~Costa, Fazel-Zarandi, Sercu, Candido, \latin{et~al.} others]{esmfold}
Lin,~Z.; Akin,~H.; Rao,~R.; Hie,~B.; Zhu,~Z.; Lu,~W.; Smetanin,~N.; dos
  Santos~Costa,~A.; Fazel-Zarandi,~M.; Sercu,~T.; Candido,~S.; others Language
  models of protein sequences at the scale of evolution enable accurate
  structure prediction. 2022\relax
\mciteBstWouldAddEndPuncttrue
\mciteSetBstMidEndSepPunct{\mcitedefaultmidpunct}
{\mcitedefaultendpunct}{\mcitedefaultseppunct}\relax
\EndOfBibitem
\bibitem[Shrake and Rupley(1973)Shrake, and Rupley]{shrake-and-rupley}
Shrake,~A.; Rupley,~J.~A. Environment and exposure to solvent of protein atoms.
  Lysozyme and insulin. \emph{Journal of molecular biology} \textbf{1973},
  \emph{79}, 351--371\relax
\mciteBstWouldAddEndPuncttrue
\mciteSetBstMidEndSepPunct{\mcitedefaultmidpunct}
{\mcitedefaultendpunct}{\mcitedefaultseppunct}\relax
\EndOfBibitem
\bibitem[Kamb and Ganguli(2025)Kamb, and Ganguli]{creativity}
Kamb,~M.; Ganguli,~S. An analytic theory of creativity in convolutional
  diffusion models. Forty-second International Conference on Machine Learning.
  2025\relax
\mciteBstWouldAddEndPuncttrue
\mciteSetBstMidEndSepPunct{\mcitedefaultmidpunct}
{\mcitedefaultendpunct}{\mcitedefaultseppunct}\relax
\EndOfBibitem
\bibitem[Shi \latin{et~al.}(2024)Shi, Han, Wang, Doucet, and Titsias]{md4}
Shi,~J.; Han,~K.; Wang,~Z.; Doucet,~A.; Titsias,~M.~K. Simplified and
  Generalized Masked Diffusion for Discrete Data. Advances in Neural
  Information Processing Systems. 2024\relax
\mciteBstWouldAddEndPuncttrue
\mciteSetBstMidEndSepPunct{\mcitedefaultmidpunct}
{\mcitedefaultendpunct}{\mcitedefaultseppunct}\relax
\EndOfBibitem
\bibitem[Nie \latin{et~al.}(2025)Nie, Zhu, You, Zhang, Ou, Hu, ZHOU, Lin, Wen,
  and Li]{llada}
Nie,~S.; Zhu,~F.; You,~Z.; Zhang,~X.; Ou,~J.; Hu,~J.; ZHOU,~J.; Lin,~Y.;
  Wen,~J.-R.; Li,~C. Large Language Diffusion Models. ICLR 2025 Workshop on
  Deep Generative Model in Machine Learning: Theory, Principle and Efficacy.
  2025\relax
\mciteBstWouldAddEndPuncttrue
\mciteSetBstMidEndSepPunct{\mcitedefaultmidpunct}
{\mcitedefaultendpunct}{\mcitedefaultseppunct}\relax
\EndOfBibitem
\bibitem[Sahoo \latin{et~al.}(2024)Sahoo, Arriola, Schiff, Gokaslan, Marroquin,
  Chiu, Rush, and Kuleshov]{simple_mdlm}
Sahoo,~S.~S.; Arriola,~M.; Schiff,~Y.; Gokaslan,~A.; Marroquin,~E.;
  Chiu,~J.~T.; Rush,~A.; Kuleshov,~V. Simple and Effective Masked Diffusion
  Language Models. Advances in Neural Information Processing Systems. 2024; pp
  130136--130184\relax
\mciteBstWouldAddEndPuncttrue
\mciteSetBstMidEndSepPunct{\mcitedefaultmidpunct}
{\mcitedefaultendpunct}{\mcitedefaultseppunct}\relax
\EndOfBibitem
\bibitem[Ou \latin{et~al.}(2024)Ou, Nie, Xue, Zhu, Sun, Li, and Li]{radd}
Ou,~J.; Nie,~S.; Xue,~K.; Zhu,~F.; Sun,~J.; Li,~Z.; Li,~C. Your Absorbing
  Discrete Diffusion Secretly Models the Conditional Distributions of Clean
  Data. 2024\relax
\mciteBstWouldAddEndPuncttrue
\mciteSetBstMidEndSepPunct{\mcitedefaultmidpunct}
{\mcitedefaultendpunct}{\mcitedefaultseppunct}\relax
\EndOfBibitem
\end{mcitethebibliography}
\end{document}